\documentclass{article}
\usepackage{arxiv}

\usepackage{color}
\usepackage{colortbl}
\usepackage{url}
\usepackage{graphicx}
\usepackage{natbib}
\usepackage[utf8]{inputenc}
\usepackage{pmboxdraw}
\usepackage{multirow}
\usepackage{afterpage}
\usepackage{longtable}

\usepackage{xcolor}
\newcommand{\comment}[1]{} % multiline comments

\title{Universal Lemmatizer: A Sequence to Sequence Model for Lemmatizing Universal Dependencies Treebanks}

\author{Jenna Kanerva, Filip Ginter and Tapio Salakoski\\
        TurkuNLP Group, Department of Future Technologies,\\
        University of Turku, Finland \\
        e-mail: jmnybl@utu.fi, figint@utu.fi, sala@utu.fi}

\begin{document}

\label{firstpage}
\maketitle

\begin{abstract}
% NATURAL LANGUAGE ENGINEERING: abstract about 300 words

In this paper we present a novel lemmatization method based on a sequence-to-sequence neural network architecture and morphosyntactic context representation. In the proposed method, our context-sensitive lemmatizer generates the lemma one character at a time based on the surface form characters and its morphosyntactic features obtained from a morphological tagger. We argue that a sliding window context representation suffers from sparseness, while in majority of cases the morphosyntactic features of a word bring enough information to resolve lemma ambiguities while keeping the context representation dense and more practical for machine learning systems. Additionally, we study two different data augmentation methods utilizing autoencoder training and morphological transducers especially beneficial for low resource languages. We evaluate our lemmatizer on 52 different languages and 76 different treebanks, showing that our system outperforms all latest baseline systems. Compared to the best overall baseline, UDPipe Future, our system outperforms it on 62 out of 76 treebanks reducing errors on average by 19\% relative. The lemmatizer together with all trained models is made available as a part of the Turku-neural-parsing-pipeline under the Apache 2.0 license.

\end{abstract}

\section{Introduction}

Lemmatization is a process of determining a base or dictionary form (lemma) for a given surface form. Traditionally, word base forms have been used as input features for various machine learning tasks such as parsing, but also find applications in text indexing, lexicographical work, keyword extraction, and numerous other language technology -enabled applications. Lemmatization is especially important for languages with rich morphology, where a strong normalization is required in applications. Main difficulties in lemmatization arise from encountering previously unseen words during inference time as well as disambiguating ambiguous surface forms which can be inflected variants of several different base forms depending on the context. 

The classical approaches to lemmatizing highly inflective languages are based on two-level morphology implemented using finite state transducers (FST)~\citep{koskenniemi1984general,karttunen1992two}. FSTs are models encoding vocabulary and string rewrite rules for analyzing an inflected word into its lemma and morphological tags. Due to surface form ambiguity the FST encodes all possible analyses for a word, and the early work on context-sensitive lemmatization was based on disambiguating the possible analyses in the given context.~\citep{smith2005context,aker2017extensible,liu2017evaluation} 

The requirement of having a pre-defined vocabulary is impractical especially when working with Internet or social media texts where the language variation is high and adaptation fast. Therefore, there has been an increasing interest in the application of context-sensitive machine learning methods that are able to deal with open vocabulary.

In this paper we present a sequence-to-sequence lemmatizer with a novel context representation. This method was used as part of the TurkuNLP submission~\citep{udst:turkunlp} in the CoNLL-18 Shared Task on Multilingual Parsing from Raw Text to Universal Dependencies~\citep{udst:overview18} where it ranked 1st out of 26 participants on the lemmatization sub-task. In addition to plain lemmatization, the system ranked 1st on the BLEX evaluation metric as well, a metric combining evaluation of both lemmatization and syntactic dependencies. Our Shared Task work is extended in several directions. Firstly, we analyze and justify the particular context representation used by the system using data from 52 languages, secondly we carry out comparison to state-of-the-art lemmatization methods, thirdly we test and evaluate two different data augmentation methods for automatically expanding training data sizes, and finally, we release the system together with models for all 52 languages as a freely available parsing pipeline, containerized using Docker for ease of use.

The rest of the paper is structured as follows. In Section~\ref{sec:ambiguity} we discuss the surface form ambiguity problem in the context of lemmatization, as well as present a data-driven study for justifying our contextual representation for resolving the problem. In Section~\ref{sec:relatedwork} we describe the most important related work. In Section~\ref{sec:methods} we present our problem setting, model architecture and implementation. Experimental setups for our main evaluation as well as results are given in Sections~\ref{sec:basic-experiments} and~\ref{sec:main-results}. In Section~\ref{sec:data-augmentation} we describe our data augmentation studies to increase training set sizes leading to a higher prediction accuracy. In Section~\ref{sec:discussion} we summarize the results as well as discuss the practical issues related to our method, most importantly prediction speed and software release. Finally we conclude the paper in Section~\ref{sec:conclusions}.

\section{Lemmatization Ambiguity and Morphosyntactic Context}
\label{sec:ambiguity}

Lemmatization methods can roughly be divided into two categories, context-aware methods where the lemmatization system is aware of the sentence context where the word appears, and methods where the system is lemmatizing individual words without contextual information. The advantage in the former approach is the ability to correctly lemmatize ambiguous words based on the contextual information while the latter is only able to either give one lemma for each word even though its lemmatization can vary in different contexts, or list all alternatives. While some of the ambiguous words are assigned the same lemma, such as \emph{love} in the verb---noun contrast (\emph{I love you} vs. \emph{Love is all you need}), are typically assigned the same lemma (\emph{love} in this case rather than \emph{to love}), it is not always the case. For example the English word \emph{lives} receives a different lemma depending on the part-of-speech (\emph{live} vs. \emph{life}). Additionally words can be ambiguous within a single part-of-speech class. For example in Finnish the word \emph{koirasta} is always a noun but depending on the grammatical case it should be lemmatized to \emph{koira} (\emph{a dog} inflected in elative case) or to \emph{koiras} (\emph{a male} inflected in partitive case). Note that the knowledge of the part-of-speech and inflectional tags, i.e.\ morphosyntactic features of the word, is sufficient to correctly lemmatize these two abovementioned examples. This holds for the majority of cases, with rare exceptions. For example, the Finnish word \emph{paikkoja} is a noun in plural partitive, but it can be an inflection of two different lemmas, \emph{paikka} (\emph{a place} or \emph{patch}) or \emph{paikko} (\emph{a spare} in bowling). In these rare cases, the meaning, and therefore the correct lemma, can only be derived from the semantic context, i.e.\ the actual meaning or topic of the sentence.

\begin{figure*}
\includegraphics[width=1.0\textwidth]{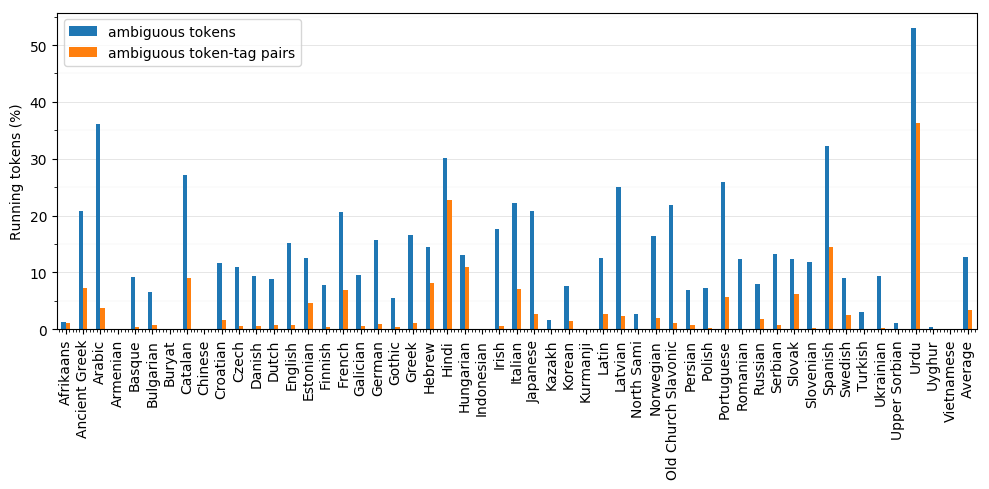}
\caption{Percentage of running tokens with ambiguous lemma and token-tag pairs with ambiguous lemma calculated from the UD v2.2 training data. An ambiguous token is a word occurring with more than one lemma in the training data, whereas an ambiguous token-tag pair is a (word, morphosyntactic tags) -tuple occurring with more than one lemma in the training data. All treebanks of one language are pooled together.}
\label{fig:amb}
\end{figure*}

\citet{bergmanis2018context} did a careful evaluation of lemmatization model effectiveness with and without contextual information. They show that including a sliding window of nearby characters significantly improves the performance compared to the context-free version of the same system. However, they only evaluate the system using a textual context (i.e.\ \emph{n} characters/words before and after the word to be lemmatized). Suspecting that this lexical context representation suffers from sparseness, we hypothesize that the morphosyntactic features will uniquely disambiguate the lemma in all but the rarest of cases, and can serve as a more practical, dense context representation for the lemmatization task. In order to establish how uniquely the features disambiguate the lemma, we measure different levels of ambiguity on the Universal Dependencies (UD) v2.2 treebanks and present the results in Figure~\ref{fig:amb}. We measure how many times a (word, morphosyntactic tags) -tuple is seen with more than one lemma compared to how many times a plain word is seen with more than one lemma in the training data.

\begin{figure*}

\includegraphics[width=0.33\textwidth]{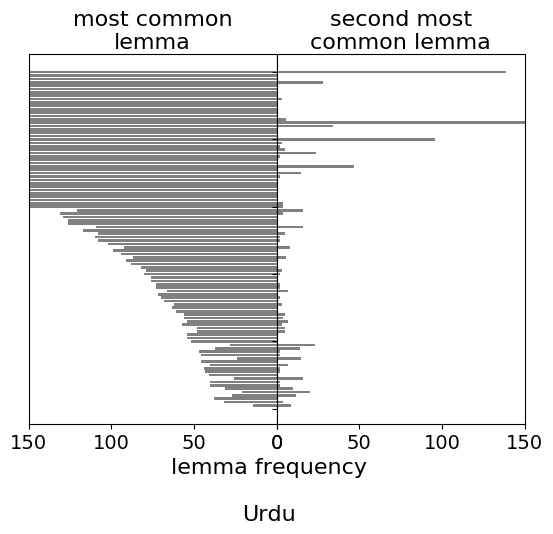}\hfill \includegraphics[width=0.33\textwidth]{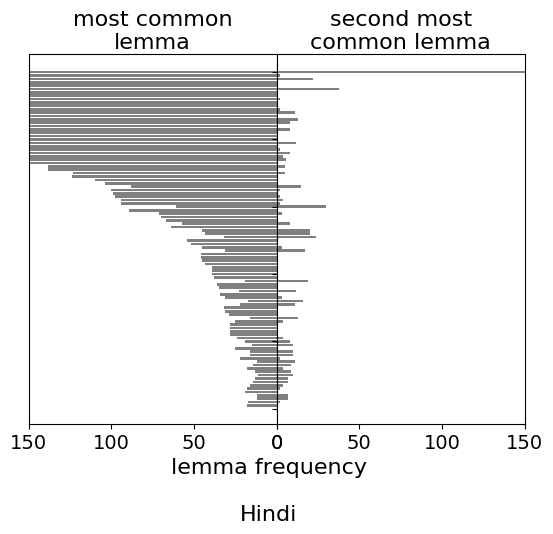}\hfill \includegraphics[width=0.33\textwidth]{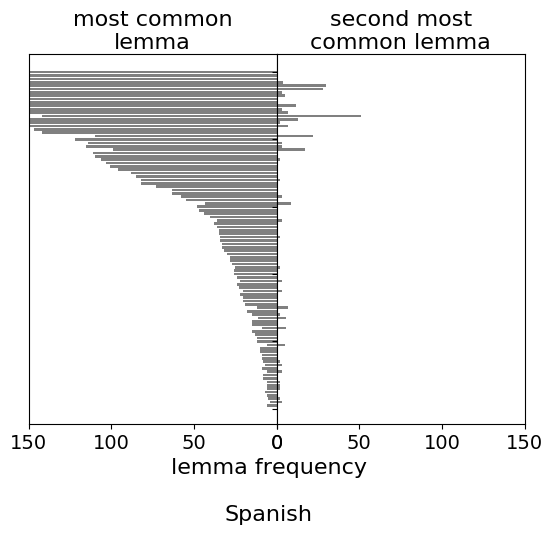}

\caption{Frequency comparison of the most common and the second most common lemmas in the training data for words which are ambiguous at the word-tag level. The top-100 most common ambiguous words are shown for Urdu (left), Hindi (middle) and Spanish (right), the three languages with the highest ambiguity rate in Figure~\ref{fig:amb}.}
\label{fig:lemma_freq}
\end{figure*}

We can see that the proportion of ambiguous lemmas drastically drops for most languages when morphosyntactic tags are taken into account, on average the token-tag pair ambiguity being close to 3\% of running tokens, while plain token ambiguity is close to 12\%. For more than half of the languages the ambiguity drops below 1\% of running tokens, to the level which does not pose an issue anymore, or, from a different point of view, can be expected to cause an issue to any machine learning system due to the rareness of the words involved as we will demonstrate shortly. However, for few languages the ambiguity remains on surprisingly high level, especially for Urdu (36\%) and Hindi (22\%), both being Indo-Aryan languages and closely related to each other, as well as for Spanish (14\%), a Romance language. To shed some light specifically on these three languages, we plot in Figure~\ref{fig:lemma_freq} the frequencies of most common and second most common lemmas for the 100 most common ambiguous words. For all three languages, and all but a handful of words, the distribution is extremely inbalanced with only a small number of occurrences of the less frequent lemma. When investigating similar cases in languages we are familiar with, we can see that in addition to real ambiguities in many cases these turn out to be annotation inconsistencies. For example, while the word \emph{vs.} as \texttt{ADP} has only one meaning in the English training data and therefore should also have only one lemma, it is lemmatized 17 times as \emph{vs.} and once as \emph{versus}. Similarly, most of the ambiguous cases in the Finnish data are inconsistencies in the placement of compound boundary markers. Even with the real ambiguities, it is debatable whether heavily skewed distributions, where the most common lemma can be several orders of magnitude more common, can be learned given the minimal number of training examples for the rarer lemma.

In the light of these findings, we therefore argue that the part-of-speech and rich morphosyntactic features are, from the practical standpoint of building a multilingual lemmatization system, sufficient to resolve the vast majority of ambiguous lemmatizations in the vast majority of the 52 languages covered by the UD dataset.

\section{Related Work}
\label{sec:relatedwork}

The most common machine learning approaches to lemmatization are based on edit tree classification, where all possible edit trees or word-to-lemma transformation rules are first gathered from the training data, and then a classifier is trained to choose the correct one for a given input word. These methods do not require that the input word is known in advance as long as the correct edit pattern is seen during training. Edit-tree classifiers are used for example in \citep{muller2015joint,straka2016udpipe,chakrabarty2017context}, and the sentence-context for resolving ambiguous words can be incorporated into these classifiers for example by using global sentence features~\citep{muller2015joint} or contextualized token representations~\citep{straka2016udpipe,chakrabarty2017context,udst:udpipefuture}.

Many recent works build on the sequence-to-sequence learning paradigm. \citet{bergmanis2018context} present the Lematus context-sensitive lemmatization system, where the model is trained to generate the lemma from a given input word one character at a time. Additionally, a context of 20 characters in each direction is concatenated with the input word, resulting in a 12\% relative error decrease compared to only the word being present in the input. The Lematus system outperforms other context-aware lemmatization systems, including \citep{chrupala2008learning,muller2015joint,chakrabarty2017context}, and can be seen at the time of writing as the current state of the art on the task. However, the task is naturally an active research area with new directions pursued e.g.\ by \citet{kondratyuk2018lemmatag}.

The 2018 CoNLL Shared Task on multilingual parsing included lemmatization as one of the objectives, and has given raise to a number of machine learning approaches. Among the top three performing systems on large treebanks, together with our work and the abovementioned edit-tree classifier of \citet{udst:udpipefuture}, ranked the Stanford system \citep{udst:stanford}. Here words whose lemma cannot be looked up in a dictionary are lemmatized using a sequence-to-sequence model without any additional context information. 

Sequence-to-sequence models have also been widely applied in the context of morphological reinflection, the reverse of the lemmatization task. In the CoNLL-SIGMORPHON 2017 Shared Task on Universal Morphological Reinflection~\citep{cotterell-EtAl:2017:K17-20} the objective was to generate the inflected word given a lemma and morphosyntactic tags. Here several of the top-ranking systems were based on sequence-to-sequence learning~\citep{kann2017lmu,bergmanis2017training}. The entry of \citet{ostling2017surug} additionally tried to boost the inflection generation by learning the primary morphological reinflection objective jointly with the reverse task of lemmatization and tagging.

\section{Methods}
\label{sec:methods}

Taking inspiration from the top systems in the CoNLL-SIGMORPHON 2017 Shared Task, we cast lemmatization as a sequence-to-sequence rewrite problem where lemma characters are generated one at a time from the given sequence of word characters and morphosyntatic tags. We diverge from previous work on lemmatization by utilizing morphosyntactic features predicted by a tagger to represent the salient information from the context, instead of using for example contextualized word representations or sliding window of text. We modify the usual order of a parsing pipeline to include the lemmatizer as the last step of the pipeline, running after the tagger and thus making it possible to access the predicted part-of-speech and morphological features at the time of lemmatization. In this study, we use the part-of-speech tagger of \citet{dozat2017stanford} modified to predict also morphological features~\citep{udst:turkunlp}. More detailed discussion of the tagger is included in Section~\ref{sec:tools}.

The input of our sequence-to-sequence lemmatizer model is the sequence of characters of the word together with the sequence of its morphosyntactic tags, while the output is the sequence of lemma characters. In the UD representation, three different columns are available for morphosyntactic tags: universal part-of-speech (\texttt{UPOS}), language-specific part-of-speech (\texttt{XPOS}) and morphological features, a sorted list of feature category and value pairs  (\texttt{FEATS}). All three are used in the input together with the word characters. For example, the input and output sequences for the English word \emph{lives} as a noun are:

\begin{verbatim}
  INPUT: l i v e s UPOS=NOUN XPOS=NNS Number=Plur
  OUTPUT: l i f e
\end{verbatim}

Once cast in this manner, essentially any of the recent popular sequence-to-sequence model architectures can be applied to the problem. Similarly to the Lematus system, we rely on an existing neural machine translation model implementation, in our case OpenNMT: Open-Source Toolkit for Neural Machine Translation~\citep{opennmt}. 

\subsection{Sequence-to-sequence Model}

\begin{figure*}

\includegraphics[width=0.9\textwidth]{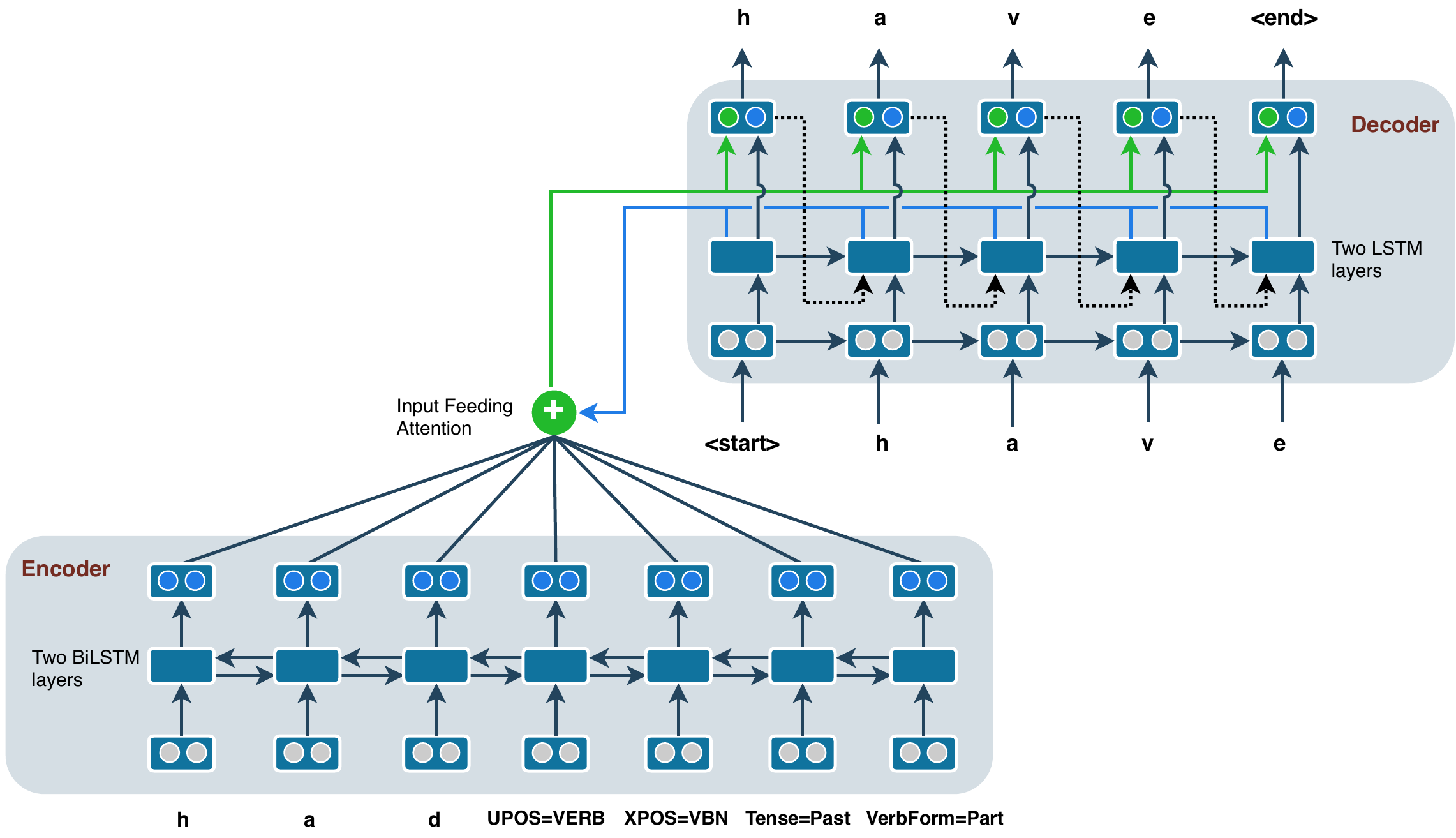}

\caption{Our encoder-decoder model architecture.}
\label{fig:network}
\end{figure*}

The model implemented by OpenNMT is a deep attentional encoder-decoder network. The encoder uses learned character and tag embeddings, and two bidirectional LSTM layers to encode the sequence of input characters and morphosyntactic tags into a same-length sequence of encoding vectors. The sequence of output characters is generated by a decoder with two unidirectional LSTM layers with input feeding attention~\citep{luong2015EMNLP} on top of the encoder output. The full model architecture is illustrated in Figure~\ref{fig:network}.

An important requirement for sequence-to-sequence models is the ability to correctly deal with out-of-vocabulary (OOV) items at inference time. For example, in machine translation foreign person and place names should often be copied into the output sequence, which is not possible if the generation is based on a straightforward classification over output vocabulary learned during training. In the case of lemmatization, this issue manifests itself as characters not seen during training. Since in some languages foreign names inflect, copying full words that contain OOV characters is not a sufficient solution. For instance, a Finnish lemmatizer model trained on a typical Finnish corpus will have a vocabulary of mostly Scandinavian characters, and will be unable to correctly lemmatize the case-inflected Czech name \emph{R\r{u}\v{z}i\v{c}kalla} into \emph{Růžička}. 

\begin{figure}
    \centering
    \includegraphics[width=0.7\textwidth]{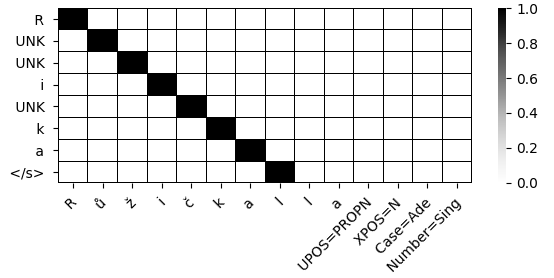}
    \caption{Visualization of the step-wise attention weights (actual system output), where the x-axis corresponds to the input sequence and the y-axis to the generated output sequence. In post-processing, each generated {\tt UNK} symbol is replaced with the input symbol that has the maximal attention at the respective timestep.}
    \label{fig:attention}
\end{figure}

In machine translation, the problem of OOV words is for the most part solved using Byte Pair Encoding (BPE) or other sub-word representations, reducing vocabulary size and handling inference-time unknown words (as unknown words can be split into known subwords)~\citep{sennrich2016neural}. As the lemmatizer operates on the level of characters, indivisible into smaller units, we instead rely on an alternative technique whereby the model is trained to predict an unknown symbol {\tt UNK} for rare and unseen characters, and as a post-processing step, each such {\tt UNK} symbol is subsequently substituted with the input symbol with the maximal attention value of the model at that point \citep{luong2015addressing,jean2015using}. For instance, for the inflected name \emph{R\r{u}\v{z}i\v{c}kalla}, we would get 

\begin{verbatim} 
INPUT: R ů ž i č k a l l a UPOS=PROPN XPOS=N Case=Ade Number=Sing
OUTPUT: R UNK UNK i UNK k a
\end{verbatim}

as the initial output of the system, later post-processed to the correct lemma \emph{Růžička} based on attention weights visualized in Figure~\ref{fig:attention}.

\section{Evaluation}
\label{sec:basic-experiments}

Next we carry out an extensive evaluation of the lemmatization framework on 52 different languages with varying lemmatization complexity and training data sizes. We compare our system to several competitive lemmatization baselines. First, we give a detailed description of our experimental setup, the baseline systems and model parameters, and after that we present the evaluation results.

\subsection{Data and Tools}

\subsubsection{Universal Dependencies Treebanks}
\label{sec:data}

We base our experiments on Universal Dependencies (UD) v2.2~\citep{udv2.2}, a multilingual collection of 122 morpho-syntactically annotated treebanks for 71 languages, with cross-linguistically consistent annotation guidelines, including also gold standard lemma annotation~\citep{nivre2016universal}. The UD treebanks therefore allow us to test the lemmatization methods across diverse language typologies and training data sizes, ranging from a little over 100 to well over 1 million tokens. We restrict the data to the subset of 82 treebanks (57 languages) used in the CoNLL-18 Shared Task on Multilingual Parsing from Raw Text to Universal Dependencies~\citep{udst:overview18}. In addition to allowing a direct comparison with the state-of-the-art parsing pipelines participating in the Shared Task, the treebanks from this subset all have a test set of at least 10,000 tokens, ensuring a reliable evaluation. Note that even though the test set is always at least 10,000 tokens, training sets may be considerably smaller, in several instances about 100 tokens.

 Furthermore, it was also necessary to remove two treebanks with no lemma annotation (Old French-SRCMF and Thai-PUD) and four treebanks with no training data (Breton-KEB, Faroese-OFT, Japanese-Modern and Naija-NSC). The four parallel ``PUD'' treebanks included in the Shared Task (Czech-PUD, English-PUD, Finnish-PUD and Swedish-PUD, each including the same 1,000 sentences translated into the target language and annotated into UD) do not have dedicated training data, but can be used as additional test sets for models trained on the Czech-PDT, English-EWT, Finnish-TDT and Swedish-Talbanken treebanks, which are sufficiently similar in annotation style. Altogether, we therefore evaluate on 76 treebanks representing 52 different languages. During evaluation we show results separately for several different groups categorizing treebanks by size or other properties. These groups are \emph{PUD} for 4 additional parallel test sets, \emph{big} for 60 treebanks with more than 10,000 tokens of training and 5,000 tokens of development data, \emph{small} for 7 treebanks with reasonably sized training data but no additional development data, and \emph{low resource} for 5 treebanks with only a tiny sample of training data (around 20 sentences) and no development data. These are the same treebank groups as defined in CoNLL-18 Shared Task.
 
 %\todo{count how many treebanks in each category: big: 76-rest, low: 5, small: 7, pud: 4}

 To ensure that treebanks in the \emph{small} and \emph{low resource} categories also have a development set for hyperparameter tuning and model selection, we adopt the data split provided by the Shared Task organizers, which creates the development set from a portion of the training data when necessary~\citep{11234/1-2859}. This data split was also used to train the Shared Task baseline model, one of the systems we compare our results to. The final numbers are always reported on the held-out test set directly specified in the UD release for each treebank. The original test section of the UD data is never used in system training and development, as suggested by the data providers and so as to be able to distribute the trained models for further comparison. For this reason we also decided not to apply N-fold cross-validation for low-resource treebanks, which otherwise would have been an option to decrease variance in the results. Furthermore, the training and development set split is also kept fixed as the development data is used only for early stopping and model selection, which we do not expect to greatly affect the numbers, and hyperparameters are not tuned separately for each treebank.

\subsubsection{Part-of-Speech and Morphological Tagger}
\label{sec:tools}

As the input of our lemmatizer is a word together with its part-of-speech and morphosyntactic features, we need a tagger to predict the required tags before the word can be lemmatized. We use the one by~\citet{udst:turkunlp}, which is based on the winning Stanford part-of-speech tagger~\citep{dozat2017deep,dozat2017stanford} from the CoNLL-17 Shared Task on multilingual parsing~\citep{udst:overview17}. The tagger has two classification layers (predicting UPOS and XPOS) over tokens in a sentence, where tokens are first embedded using a sum of learned, pre-trained and character-based LSTM embeddings, which are then encoded with a bidirectional LSTM to create a sequence of contextualized token representations. The classification layers are trained jointly on top of these shared token representations. By default, the original tagger does not predict the rich morphosyntactic features (FEATS column in CoNLL-U format). To this end, in~\citet{udst:turkunlp} we modified the tagger training data by concatenating the morphosyntactic features with the language-specific part-of-speech tag (XPOS), thereby forcing the tagger to predict the XPOS tag and all morphosyntactic features as one multi-class classification problem. For example, in Finnish-TDT the original XPOS value \texttt{N} and FEATS value \texttt{Case=Nom|Number=Sing} are concatenated into one long string \texttt{XPOS=N|Case=Nom|Number=Sing} which is then predicted by the tagger. The morphological features are sorted so as to avoid duplicating label strings having the same tags in different order. After prediction, the morphosyntactic features are extracted into a separate column. The evaluation in~\citet{udst:turkunlp} shows that this data manipulation technique does not harm the prediction of the original XPOS tag, and accuracy of morphosyntactic feature prediction (FEATS field) is comparable to the state-of-the art in the CoNLL-18 Shared Task, ranking 2nd in the evaluation metric combining both morphosyntactic features and syntactic dependencies, and 3rd in the evaluation of plain morphosyntactic features. In our preliminary experiments, we expected the complex morphology of some languages to result in a large number of very rare feature strings if combined in such a simple manner. We tested several models, for instance predicting a value for each category separately (for example \emph{Nominative} for \emph{Case}) from a shared representation layer. However, the results were surpassed by the simple concatenation of morphological features. The conclusion of this experiment was that even though some languages have many unique feature combinations (number of unique combinations ranging from 15 to ~2,500) the most common ones cover the vast majority of the data, with the rare classes having no practical effect on the prediction accuracy (more detailed discussion is given in ~\citet{udst:turkunlp}).

\subsection{Parameter Optimization}

\begin{figure*}

\includegraphics[width=0.49\textwidth]{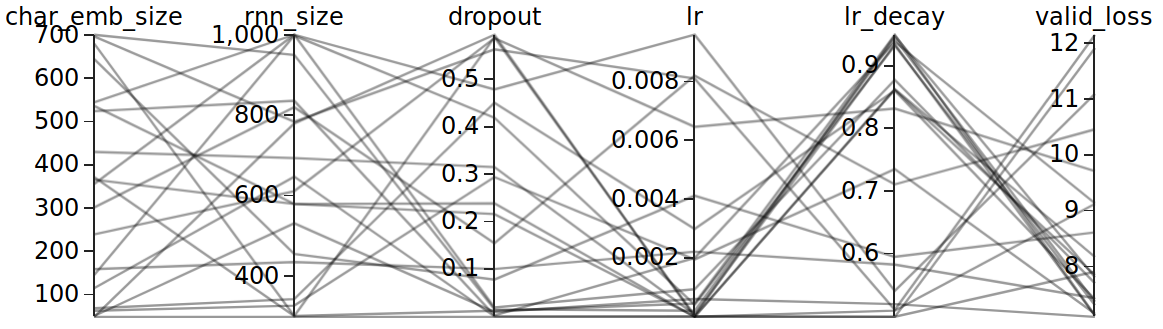}\hfill \includegraphics[width=0.49\textwidth]{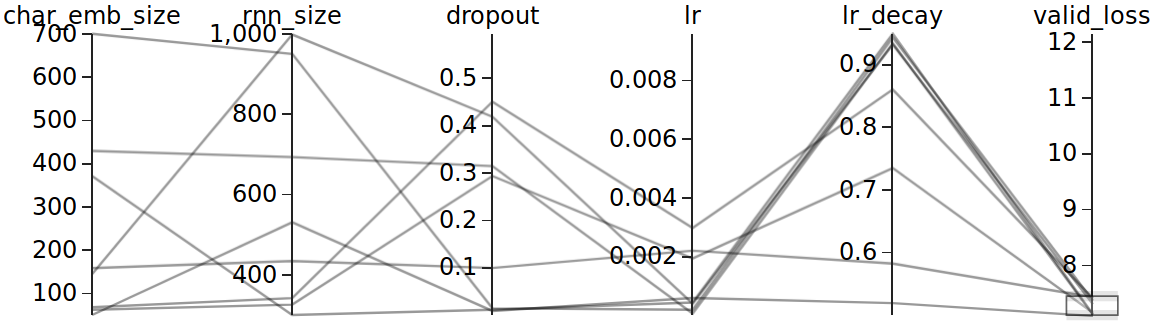}
\includegraphics[width=0.49\textwidth]{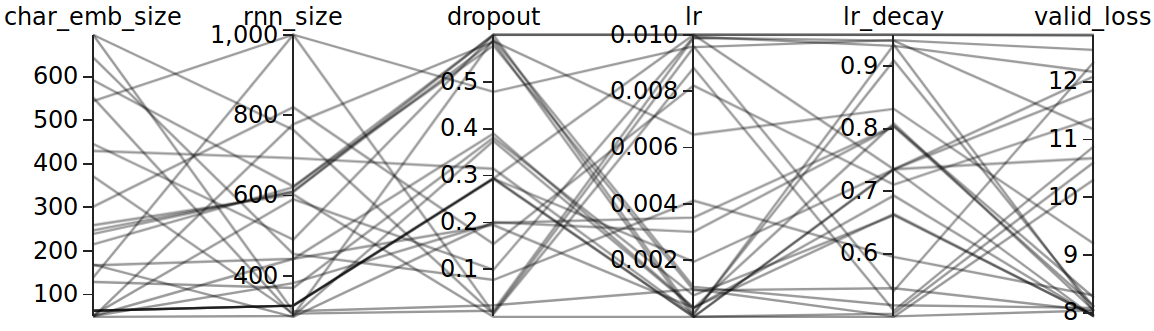}\hfill \includegraphics[width=0.49\textwidth]{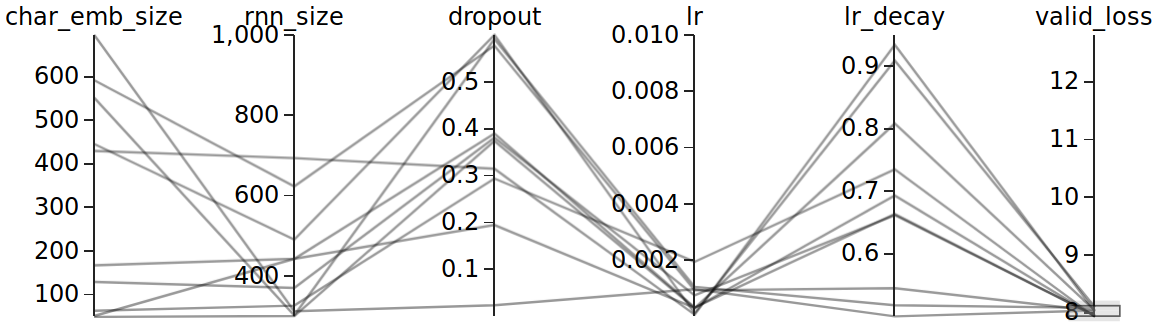}
\includegraphics[width=0.49\textwidth]{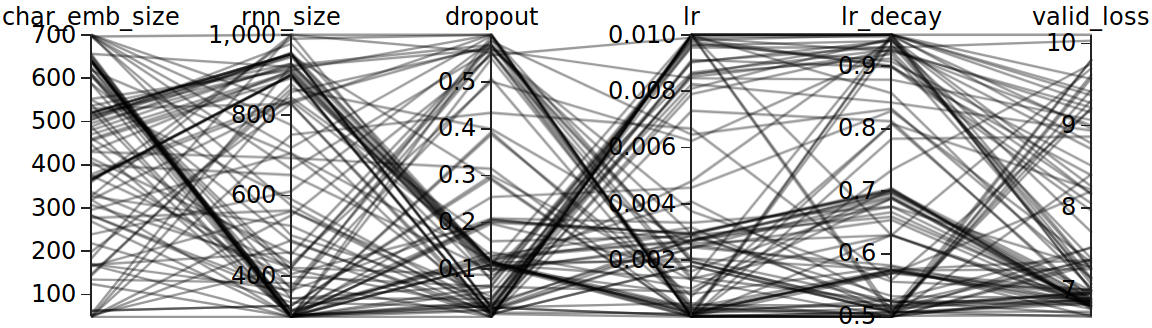}\hfill \includegraphics[width=0.49\textwidth]{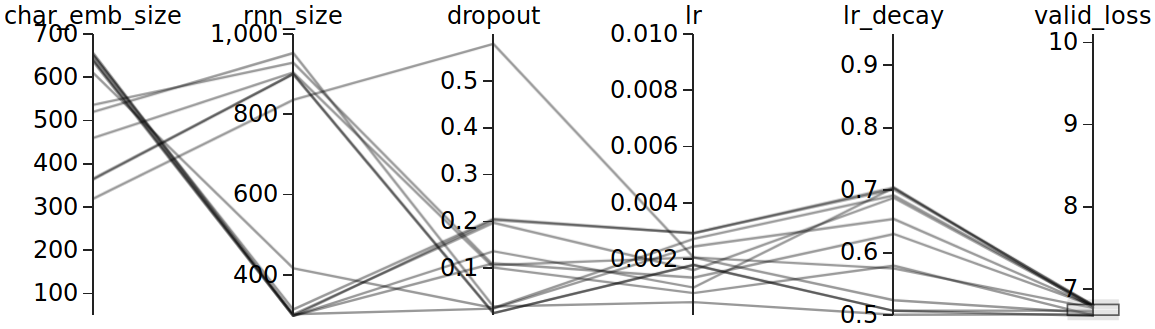}

\caption{Parallel coordinates graphs for visualizing hyperparameter optimizer runs for three different treebanks (top: English, middle: Finnish, bottom: Irish). On the left side of the figure are all optimizer runs completed during the 24 hour time window, while on the right side these runs are filtered based on the validation loss to demonstrate parameter ranges resulting in low validation loss values.}
\label{fig:optimizer}
\end{figure*}

To optimize the hyperparameters of our lemmatization models, we use the \mbox{RBFOpt} library designed for optimizing complex black-box functions with costly evaluation~\citep{costa2014rbfopt}. Different values of embedding size, recurrent layer size, dropout, learning rate, and learning rate decay parameters are experimented with. We let the RBFOpt optimizer run for 24 hours on three different treebanks, completing about 30 training runs for Finnish and English, and about 300 for the much smaller Irish treebank. The findings are visualized in Figure~\ref{fig:optimizer}: On the left side of the figure all different runs completed by the optimizer are shown as a parallel coordinates graph, while on the right side we use a validation loss filter to show only those runs that result in low validation loss values. From this we can more easily determine the optimal parameter ranges and their mutual relationship.

Based on these optimizer runs, the lemmatization models seem to be moderately stable, most of the parameter values having individually only a small influence on the resulting validation loss, once the RBFOpt optimizer finds the appropriate region in the parameter space. The learning rate parameter ({\tt lr} column) appears to have the largest impact, where lower learning rate values generally work better. Overall, the learning is stable across the parameter space, and the parameter optimization does not play a substantial role. Even default values as defined in the OpenNMT toolkit worked comparatively well. 

In the final experiments, apart from the batch size, uniform hyperparameter settings based on the observations of the three optimization runs are used for all treebanks. We set the embedding size to 500, dropout to 0.3, recurrent size to 500, and we use the Adam optimizer~\citep{kingma2015adam} with initial learning rate of 0.0005 and learning rate decay with 0.9 starting after 20 epochs. All models are trained for 50 epochs, but for smaller treebanks we decrease the minibatch size to increase the number of updates applied during training. Our default minibatch size is 64, but for treebanks with less than 2,000 training sentences and less than 200 training sentences we use 32 and 6, respectively. Models usually converge around epochs 30--40, and final models are chosen based on prediction accuracy on the validation set. During prediction time we use beam search with beam size 5.

\subsection{Baselines}
\label{sec:baselines}

We compare our lemmatization performance to several, recent baseline systems. \emph{Baseline UDPipe}~\citep{straka2016udpipe} is the organizers' baseline parsing pipeline from the CoNLL-18 Shared Task, which, due to its easy usability and availability of pretrained models, has been the go-to tool for parsing UD data. \emph{UDPipe Future}~\citep{udst:udpipefuture} is an updated version of the baseline UDPipe pipeline ranking high across the CoNLL-18 ST evaluation metrics. Both UDPipe versions have a lemmatizer based on the edit-tree classification method. The \emph{Stanford} system~\citep{udst:stanford} is a dictionary look-up followed by a context-free sequence-to-sequence lemmatizer for words unseen in the training data. Together with our entry, \emph{UDPipe Future} and \emph{Stanford} form the top three performing entries in the lemmatization evaluation of the CoNLL-18 ST on the big treebank category. In addition to top ranking systems from the CoNLL-18 ST, we also compare to the context-aware \emph{Lematus} sequence-to-sequence lemmatizer~\citep{bergmanis2018context} which outperformed all its baselines in the earlier studies, and can be seen as a current state-of-the-art in lemmatization research. Our final baseline (\emph{Look-up}) is a simple look-up table, where lemmas are assigned based on the most common lemma seen in the training data, while unknown words are simply copied unchanged to the lemma field.

Results for the baseline systems from the CoNLL-18 ST (\emph{Baseline UDPipe}, \emph{UDPipe Future} and \emph{Stanford}) are obtained directly from the official ST evaluation\footnote{Evaluation results are available at \url{http://universaldependencies.org/conll18/results-lemmas.html}.}, while the \emph{Lematus} models are reimplemented using the OpenNMT toolkit to overcome the experimental differences between this and the original study and performance issues regarding the original implementation\footnote{The original implementation relies on the outdated Theano backend which is no longer compatible with our GPU servers.}. To mimic the CoNLL-18 ST lemmatization evaluation settings, where lemmas are evaluated on top of the predicted sentence and word segmentation, we apply the segmentation of the \emph{Baseline UDPipe} system~\citep{11234/1-2859} for our lemmatizer as well as for the \emph{Lematus} and \emph{Look-up} baselines. The \emph{UDPipe Future} and \emph{Stanford} systems instead have their own built-in segmenters. However, \citet{udst:udpipefuture} reports that when using the same segmentation as in our pipeline, the lemmatization accuracy of \emph{UDPipe Future} decreased by 0.03pp overall, showing that the difference between our and \emph{UDPipe Future} segmentation is not significant. For the \emph{Stanford} system comparable numbers are not available, and we need to rely on the official Shared Task evaluation.\footnote{Note that the Stanford system official results are affected by a known segmentation bug. Overall lemmatization results reported by the Stanford team for their corrected system improve its performance from -2.92pp to -2.07pp difference to our system, i.e.\ not affecting the overall conclusions.}

\section{Results}
\label{sec:main-results}

%\todo{count how many treebanks in each category: big: 76-rest, low: 5, small: 7, pud: 4}

\begin{figure*}
\includegraphics[width=0.95\textwidth]{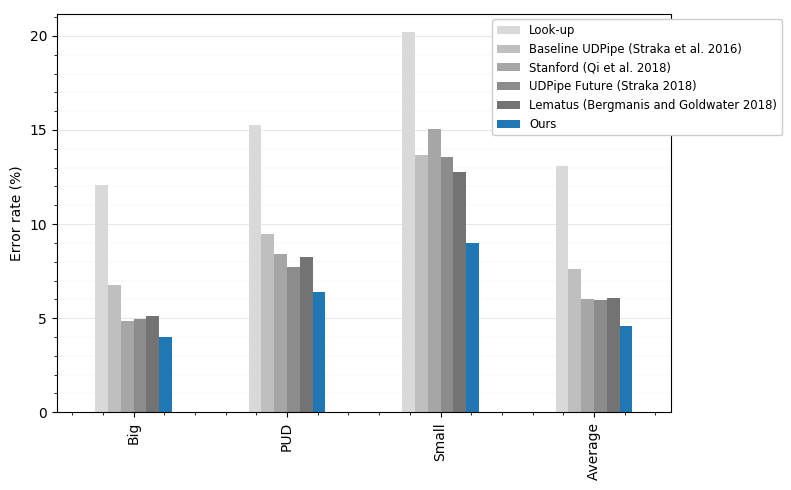}
\caption{Test set word-level error rates for our system as well as all baseline systems divided into three different treebank groups, big, PUD and small, as well as macro-average over all treebanks belonging to these groups.}
\label{fig:main_results}
\end{figure*}

The results are shown in Figure~\ref{fig:main_results}, where we measure word-level error rates separately on three treebank categories, \emph{big}, \emph{PUD} and \emph{small}, as well as macro-average error rate over all treebanks belonging to these three categories.

On all three categories our system outperforms all the baselines with an overall error rate of 4.61 (macro-average across the treebanks in the three categories). Compared to the second best overall system, \emph{UDPipe Future}, our error rate is 1.35 absolute percent point lower, reducing errors by 23\% within these three treebank categories. The widest margin from our system to the second best systems is in the \emph{small} treebank category where our system reduces errors by 30\%, from 12.75 to 8.98, compared to the second best \emph{Lematus} system. The simplistic \emph{Look-up} baseline is clearly worse than all other systems, reflecting that plain memorizing training tokens and fallback copying unknowns is not a sufficient strategy for language universal lemmatizer. The three most recent baseline systems (\emph{Stanford}, \emph{UDPipe Future} and \emph{Lematus}) perform evenly in terms of average error rate, outperforming the older \emph{Baseline UDPipe}.

% ERROR RATES
%                   Big    PUD  Small  Average
%Baseline UDPipe   6.77   9.48  13.66     7.60
%Lematus           5.14   8.25  12.75     6.06
%Look-up          12.09  15.24  20.18    13.07
%Ours              3.98   6.41   8.98     4.61
%Stanford          4.83   8.39  15.07     6.04
%UDPipe Future     4.96   7.70  13.54     5.96

\begin{figure*}
\includegraphics[width=0.8\textwidth]{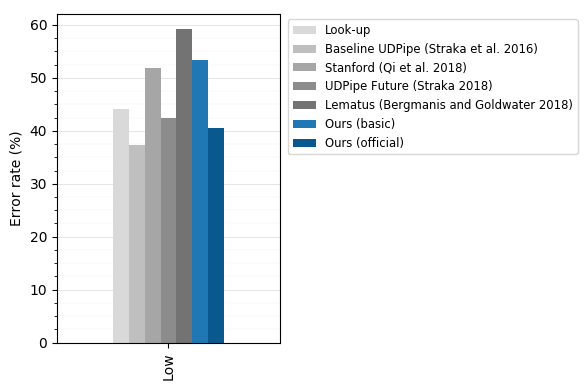}
\caption{Test set macro-average error rates of five low-resource category treebanks for two our models as well as all baseline systems.}
\label{fig:low_resource}
\end{figure*}

The fourth treebank category used in the CoNLL-18 ST is \emph{low-resource}, where only a tiny training data sample is available, usually around 20 sentences. Results for this group are given separately in Figure~\ref{fig:low_resource} where we measure macro-average word-level error rate over the five treebanks belonging to this category. Few dozens of training sentences cannot be expected to result in a well-performing lemmatization system, and indeed, all systems have error rates near 40--50\%, where almost half of the tokens are lemmatized incorrectly. Here even the \emph{Look-up} baseline performs comparably to the other systems, which is for the most part caused by the fallback copying of the unknown words unchanged to the lemma field, and therefore getting the easy words correct. For our system, we report two different runs, \emph{basic} is trained purely on the tiny training data sample, while \emph{official} is our official submission for the CoNLL-18 ST where we experimented with preliminary data augmentation methods for automatically enriching the tiny training data sample with words analyzed by morphological transducers. The two lowest average error rates in the low-resource category are achieved by the two different versions of UDPipe (UDPipe Baseline and UDPipe Future), both belonging to the category of edit-tree classification systems. Systems based on sequence-to-sequence learning (Stanford, Lematus and ours) are hypothesized to be more data hungry, and these systems indeed achieve clearly worse results in the low-resource category, all making more errors than correct predictions. However, when we include the additional training data obtained with data augmentation methods, we are able to boost our performance (\emph{Our official}) to the level of the two edit-tree classification systems reducing errors by 24\% compared to our basic models. Nevertheless, as all results are about the same level as the simple \emph{Look-up} baseline, the achieved improvement is mostly theoretical.

% ERROR RATES      Low
%Baseline UDPipe  37.32
%Lematus          59.19
%Look-up          44.15
%Ours (basic)     53.36
%Ours (official)  40.52
%Stanford         51.79
%UDPipe Future    42.47

\section{Training Data Augmentation} 
\label{sec:data-augmentation}

In our initial attempt to improve lemmatization performance on the low-resource languages in the CoNLL-18 Shared Task, we observed a substantial improvement over our basic run when the morphological transducers are used to generate additional training data. However, the overall accuracy of those datasets is below the limits of usable real-world systems and thus the seen improvements are more theoretical than practical. Next, we investigate whether automatic training data augmentation methods are useful for languages with much better baseline accuracy to improve lemmatization performance in a real-life setting as well. We test two different methods on a full set of treebanks suitable for a given method. First, we apply an autoencoder style secondary learning objective, where the lemmatizer model is trained to repeat the given input sequence without any modification. The benefit of such objective is to support the stem generation without requiring any additional resources. Secondly, we repeat the experiment with the morphological transducers for all languages which have an Apertium morphological transducer available. We generate additional inflection--lemma pairs based on the known vocabulary and inflection paradigms encoded as a transducer, and these new training examples are then mixed with the original training data. Next, we explain both data augmentation methods in detail, and afterwards compare the results.

\subsection{Autoencoding Random Strings}

In our first data augmentation method we apply joint learning of autoencoding and lemmatization. The basis of the required work in sequence-to-sequence lemmatization is the ability to repeat the word stem in the output generation. As suggested by \citet{kann2017unlabeled} in the context of morphological reinflection we hypothesize that learning to repeat the input characters as a secondary task with additional training examples could simplify the lemmatization complexity the model has to learn especially for treebanks with less training data. If the model is taught separately to repeat the input characters in the generated output, the actual lemmatization rewriting task could be learnable with less training material. In particular, this approach should be able to help in low-resource settings when the amount of training data is not necessarily sufficient for learning the complex task from scratch.

Following the autoencoding idea of \citet{kann2017unlabeled}, we enrich our lemmatization training data for each treebank by adding randomly generated strings where the input and output sequences are verbatim copies. These random strings are not equipped with any morphosyntactic tags, but instead a special tag is added to give the model the ability to distinguish these from the actual lemmatization examples to avoid confusion. Each random string is generated by sampling with replacement 3-12 characters individually from the known character vocabulary with character probabilities calculated from the training data, producing word-like items of varying lengths. However, we force each character in the vocabulary to be sampled at least once to better cover the known character vocabulary. This is achieved by first generating as many random strings as there are characters in the alphabet, each string containing the respective alphabet character at a random position. The rest of the strings are randomly sampled without any further restrictions on the alphabet. These generated strings are then mixed together with the actual training examples by randomly shuffling all training examples, and both tasks are thus trained simultaneously. The random shuffling of training examples (i.e.\ individual words), and therefore breaking the semantic context, does not harm the training of our lemmatizer as it is anyway looking at individual words at a time. As in our training data the morphosyntactic tags are already included for each word, and the random autoencoder strings do not use any morphosyntactic tags, there is no requirement of running the tagger at training time, thus making the training data shuffling procedure straightforward. We chose to autoencode random strings rather than actual words as that way we do not need any external resources and the method is easily repeatable for any language.

\subsection{Morphological Transducers}

In our second data augmentation method we lean on additional morphological/lexical resources available for a particular language. In addition to Universal Dependencies, other projects are also striving to build unified morphological resources across many different languages. For example the UniMorph project~\citep{unimorph2016} extracts and normalizes morphological paradigms from the Wiktionary free online dictionary site. Further, finite state transducers for morphological analysis and generation for a multitude of languages are available in the Apertium framework, which includes a pool of open source resources for natural language processing~\citep{tyers2010free}. Both UniMorph and Apertium frameworks can be used to collect inflected words and for each word a set of possible lemmas together with the corresponding morphological features. However, while these resources are unified within a project, their schema and guidelines differ from each other across different projects. For this reason using a mixture of training examples gathered from two or more different sources is not a straightforward task. While harmonized annotations across different languages give a good starting point for multilingual conversion, the mapping is usually not fully deterministic (see e.g.~\citet{mccarthy2018marrying} for detailed study of mapping from Universal Dependencies into UniMorph).

We expand our preliminary data augmentation experiments carried out during the CoNLL-18 ST where we used the Apertium morphological transducers to collect additional training examples. A morphological transducer is a finite-state automaton including morphological paradigms (inflection regularities/rules) and a lexicographical database (lexicon), where each lexical entry (lemma) is assigned to the inflection paradigm it follows. These linguistic resources can be compiled into an efficient finite-state transducer, an automaton which is able to return all matching lemmas and morphological hypotheses encoded in it for the given input word.

We set out to test whether improvements similar to those achieved with low-resource languages can also be seen with languages already including a reasonable amount of initial training data. We develop a language-agnostic feature mapping from Apertium features into UD, allowing us to cover all UD languages which have an Apertium morphological transducer available (Arabic, Armenian, Basque, Bulgarian, Buryat, Catalan, Czech, Danish, Dutch, English, Finnish, French, Galician, German, Greek, Hindi, Italian, Kazakh, Kurmanji, Latvian, Norwegian, Polish, Russian, Spanish, Swedish, Turkish, Ukrainian and Urdu). 

For each of these languages we first gather a full vocabulary list sorted by word frequencies in descending order. These lists are gathered mainly from the web crawl datasets~\citep{conll-raw-data}, but for languages not included in the distributed web crawl dataset (Armenian, Buryat, Kurmanji) we use Wikipedia dumps instead. The word frequency lists are then analyzed by the Apertium morphological transducers where for each unique word we obtain a set of possible lemmas and their corresponding morphological features. Words not recognized by the transducer (not part of the predefined lexicon) are simply discarded. All of these Apertium analyzes are then converted into the UD schema using our language-agnostic feature mapping where each morphological feature is converted into UD, based on a manually created look-up table. As the mapping from Apertium features into UD features is not a fully deterministic task, our language-agnostic feature mapping is designed for high precision and low recall, meaning that if a feature cannot be reliably translated, it will be dropped from the UD analyses. This approach may produce incomplete UD analyses, but we hypothesize that the lemmatizer model is robust enough to be able to utilize existing features without missing ones being too harmful for the training process, especially since in the actual training data these augmented examples are mixed together with the actual ones. The lemmas, on the other hand, we assume to be relatively harmonized between UD and Apertium by default, and these are used without any conversion or modification. After feature translation, we skip words which already appear in the original treebank training data, as well as all lemmas with a missing part-of-speech tag in the UD analysis due to an incomplete feature conversion, and all ambiguous words having two or more different lemmas with exactly the same morphological features. Finally, we pick a number of most common words from the UD converted and filtered transducer output, which are then mixed together with the original treebank training data. All training examples are randomly shuffled before training.

% ONLY TRANSDUCER TREEBANKS
% Our basic              92.033191
% Our autoencoder        93.105319
% Our transducer         93.545745
% Our mixed 2K+2K        93.454043
% Our mixed 4K+4K        93.564894
% Our mixed 8K+8K        93.612979
% UDPipe Future          92.163404
% Transducer recall      78.146809
% Transducer Coverage    86.760638

% ALL TREEBANKS (if no transducer, basic is copied)
% Our basic              92.216842
% Our autoencoder        92.889079
% Our transducer         93.152237
% Our mixed 2K+2K        93.119474
% Our mixed 4K+4K        93.173289
% Our mixed 8K+8K        93.237500
% UDPipe Future          91.639737

% ALL TREEBANKS EXCEPT LOW (if no transducer, basic is copied)
% Our basic              95.370704
% Our autoencoder        95.423944
% Our transducer         95.446056
% Our mixed 2K+2K        95.472535
% Our mixed 4K+4K        95.478169
% Our mixed 8K+8K        95.508873
% UDPipe Future          94.042113

\begin{table*}
\centering
\begin{tabular}{lccc}
\hline
\hline
\multirow{ 2}{*}{\textbf{Model}}       &  \textbf{All} &  \textbf{Excl. low}     & \textbf{Transd. only} \\
                       & \textbf{treebanks}  & \textbf{resource} &    \textbf{treebanks} \\\hline
Basic                  & 92.22          & 95.37    & 92.03 \\
Augm. autoencoder 4K   & 92.89          & 95.42    & 93.11 \\
Augm. transducer 4K    & 93.15          & 95.45    & 93.55 \\
Augm. mixed 2K + 2K    & 93.12          & 95.47    & 93.45 \\
Augm. mixed 4K + 4K    & 93.17          & 95.48    & 93.56 \\
Augm. mixed 8K + 8K    & \textbf{93.24}          & \textbf{95.51}    & \textbf{93.61} \\
Transd. Coverage       & ---            &  ---     & 86.76 \\
Transd. Recall         & ---            &  ---     & 78.15 \\\hline\hline

\end{tabular}
\caption{Evaluation of our two data augmentation methods, augmented with autoencoder and augmented with transducer as well as a mixed method, compared to our basic models. Additionally, we measure average percentage of words recognized by the transducer (Transducer Coverage) and average percentage of words having the correct lemma among the possible analyses (Transducer Recall), which represents an oracle accuracy achievable by transducers if all lemmas could be disambiguated correctly. All metrics are measured on token level.}
\label{tbl:aug_results}
\end{table*}

\subsection{Data Augmentation Results}

First we compare the two augmentation methods against our basic system, where, based on observations in \citet{bergmanis2017training}, we mix 4,000 additional training examples together with the original training data in both experiments. We decided to use a constant number of additional examples rather than a percentage to better account for the low-resource languages, the ones benefiting most from the experiment, where for example a 20\% increase in training data would still translate to having less than 500 training examples. Secondly, we add experiments on using a mixture of both augmentation techniques and increasing the number of additional examples included. Additionally, we test how well a morphological transducer itself could serve as a lemmatizer by measuring its coverage (how many words from the test data are recognized by the transducer) and lemma recall (how many words from the test set have the correct lemma among the possible analyses given by the transducer). Lemma recall therefore gives an upper-bound, oracle accuracy achievable by the transducer, assuming that all lemmas in its output can be correctly disambiguated. Results are given in Table~\ref{tbl:aug_results}. We measure macro accuracy over all treebanks and results are given separately for three treebank groups: \emph{All treebanks} includes all 76 treebanks studied in this paper, \emph{Excluding low resource} is all treebanks except the five low resource treebanks, and \emph{Transducer-only treebanks} is a set of 47 treebanks representing languages which have a morphological transducer available. Note that in \emph{All treebanks} results the \emph{Augm. transducer} row uses the basic model for treebanks where a transducer is not available, giving a realistic comparison against the \emph{Augm. autoencoder} method which does not suffer from lacking resources. In the mixed experiments, if a transducer is not available for a language, the training data is enriched only with the autoencoder examples. The two direct transducer metrics (Transducer Coverage and Recall), however, can be realistically measured only for languages having a transducer available and the results reported for the \emph{Transducer-only treebanks} group allow for a direct comparison between plain transducers and our models.

% all treebanks: 1 - (100 - 93.24) / (100 - 92.22) = 13.11 = 13 %
% excluding low: 1 - (100 - 95.51) / (100 - 95.37) = 3.02 = 3 %

In all three groups, all augmentation methods are able to surpass the basic model, with the transducer-based method giving slightly better overall results than the autoencoder. When mixing the two methods, the same amount of total examples as in the plain transducer augmentation is divided evenly between the two methods. The mixed method is not able to surpass the transducer-based one, but when increasing the amount of additional mixed data, the performance also increases slightly, the mixed 8K + 8K, the largest mixed method tested, giving the best overall performance. When considering a macro-average over all treebanks, errors are reduced by 13\% relative compared to our basic models. However, when excluding the five low resource treebanks already discussed in Section~\ref{sec:main-results} the difference is smaller, and the relative error reduction becomes a mere 3\%, demonstrating that --- unsurprisingly --- most of the benefit comes from the low resource languages and only a minimal improvement can be seen with reasonably-sized training data sets.

The average coverage for the morphological transducers is ~86\%, with recall being ~78\%. These numbers are clearly below our lemmatization methods, showing that, averaged across many languages, the approach relying on a pre-defined lexicon and ruleset does not fare favorably to sequence-to-sequence machine learning methods. The average transducer coverage is on par with the one reported by \citet{tyers2010free}, where coverage numbers reported for a set of languages varies between 80\% and 98\%, however with our set of languages the variation is much higher ranging between 5\% and 99\%, and clearly the transducers in the lower coverage region are missing much of the core vocabulary. These are measured without using morphological guessers, where unknown words can be analysed based only on their morphological shape (for example known suffixes). However, as the guessers consider every possible mapping allowed by the rules of the language, in many cases a great number of different alternatives is returned, which would need to be disambiguated later on. We therefore leave it as a future work to study whether morphological guessers and sequence-to-sequence lemmatizers can have a shared interest. By comparing the transducer coverage and recall, we can have an estimate of how harmonized the lemmas are between Apertium transducers and UD treebanks on average. If 86\% of words are recognized by the transducer, but only 78\% are having a ``correct'' lemma analysis, then 8\% of the treebank words are recognized but with a ``wrong'' lemma, hinting at an incompatible analysis. We leave it as a future study to examine, whether the differences are systematic and further gains could be obtained with filtering or harmonizing the lemma annotations between Apertium and UD in addition to harmonizing morphological features. Such a study however requires the knowledge of each of the involved languages.

\section{Discussion}
\label{sec:discussion}

\subsection{Result Summary}

In Table~\ref{tbl:big_table} we summarize the results of all the major experiments reported in this paper. For each treebank we present the accuracy of our best overall method, \emph{Augm. mixed 8K + 8K}, and for comparison, we also add results for our basic method as well as the best overall baseline method, \emph{UDPipe Future}. The comparison of our system and the UDPipe Future baseline is visualized by coloring each line green where our Mixed 8K+8K method is better than the UDPipe Future baseline. As discussed in Section~\ref{sec:baselines}, all numbers are measured on top of predicted segmentation, therefore reflecting a realistic expectation of the performance with no gold-standard data used at any point during prediction.

Out of the 76 treebanks, our method outperforms the UDPipe Future baseline on 62 treebanks. On average, across the 76 treebanks, this translates to a relative 19\% error reduction. On 36 treebanks the relative error reduction is more than 20\%, meaning that we are able to remove at least one fifth of the errors the best baseline system is making.

\definecolor{lightgreen}{rgb}{0.9,1,0.9}

%%% MACHINE GENERATED TABLE %%%
\afterpage{
{\small
\begin{longtable}{lllrrrl}

%\begin{tabular}{lllrrrl}

\hline\hline
\multirow{ 2}{*}{\textbf{Treebank}} & \textbf{Treebank}  &  \textbf{UDPipe} &  \textbf{Our}  &  \textbf{Our augm.} & \textbf{Relative diff} \\
                                    & \textbf{category}  &  \textbf{Future} & \textbf{basic} & \textbf{mixed} & \textbf{UDP-Ours} \\\hline \endhead
 \rowcolor{lightgreen}         Afrikaans-AfriBooms &               big &          97.11 &      97.59 &            97.76 &                      22.5\% \\
 \rowcolor{lightgreen}         Ancient Greek-PROIEL &               big &          91.08 &      97.27 &            97.31 &                      69.8\% \\
 \rowcolor{lightgreen}        Ancient Greek-Perseus &               big &          81.78 &      89.40 &            89.60 &                      42.9\% \\
 \rowcolor{lightgreen}                  Arabic-PADT &               big &          88.94 &      89.47 &            89.46 &                       4.7\% \\
 \rowcolor{lightgreen}              Armenian-ArmTDP &               low &          57.46 &      66.82 &            71.81 &                      33.7\% \\
 \rowcolor{lightgreen}                   Basque-BDT &               big &          95.19 &      96.66 &            96.81 &                      33.7\% \\
 \rowcolor{lightgreen}                Bulgarian-BTB &               big &          97.41 &      98.21 &            98.17 &                      29.3\% \\
                                         Buryat-BDT &               low &          56.83 &      25.55 &            56.05 &                      -1.8\% \\
                                     Catalan-AnCora &               big &          98.90 &      97.57 &            97.66 &                     -53.0\% \\
                                        Chinese-GSD &               big &          90.01 &      87.74 &            89.55 &                      -4.4\% \\
 \rowcolor{lightgreen}                 Croatian-SET &               big &          96.69 &      96.81 &            96.87 &                       5.4\% \\
 \rowcolor{lightgreen}                    Czech-CAC &               big &          98.14 &      98.19 &            98.34 &                      10.8\% \\
 \rowcolor{lightgreen}                Czech-FicTree &               big &          97.80 &      98.74 &            98.84 &                      47.3\% \\
                                          Czech-PDT &               big &          98.71 &      98.48 &            98.52 &                     -12.8\% \\
                                          Czech-PUD &               PUD &          96.44 &      96.04 &            96.14 &                      -7.8\% \\
 \rowcolor{lightgreen}                   Danish-DDT &               big &          96.66 &      97.79 &            97.88 &                      36.5\% \\
 \rowcolor{lightgreen}                 Dutch-Alpino &               big &          96.76 &      96.67 &            96.85 &                       2.8\% \\
 \rowcolor{lightgreen}             Dutch-LassySmall &               big &          95.78 &      97.40 &            97.44 &                      39.3\% \\
                                        English-EWT &               big &          97.23 &      96.96 &            96.94 &                      -9.5\% \\
 \rowcolor{lightgreen}                  English-GUM &               big &          96.18 &      96.07 &            96.21 &                       0.8\% \\
 \rowcolor{lightgreen}                English-LinES &               big &          96.44 &      96.54 &            96.79 &                       9.8\% \\
 \rowcolor{lightgreen}                  English-PUD &               PUD &          95.87 &      96.39 &            96.40 &                      12.8\% \\
 \rowcolor{lightgreen}                 Estonian-EDT &               big &          94.88 &      96.56 &            96.60 &                      33.6\% \\
 \rowcolor{lightgreen}                  Finnish-FTB &               big &          94.74 &      97.02 &            97.18 &                      46.4\% \\
 \rowcolor{lightgreen}                  Finnish-PUD &               PUD &          90.64 &      95.05 &            95.13 &                      48.0\% \\
 \rowcolor{lightgreen}                  Finnish-TDT &               big &          90.18 &      95.24 &            95.40 &                      53.2\% \\
 \rowcolor{lightgreen}                   French-GSD &               big &          96.75 &      96.90 &            96.91 &                       4.9\% \\
 \rowcolor{lightgreen}               French-Sequoia &               big &          97.36 &      97.98 &            98.06 &                      26.5\% \\
 \rowcolor{lightgreen}                French-Spoken &               big &          95.98 &      96.77 &            97.04 &                      26.4\% \\
 \rowcolor{lightgreen}                 Galician-CTG &               big &          97.53 &      97.88 &            97.92 &                      15.8\% \\
 \rowcolor{lightgreen}             Galician-TreeGal &             small &          95.05 &      94.98 &            95.50 &                       9.1\% \\
 \rowcolor{lightgreen}                   German-GSD &               big &          96.14 &      96.68 &            96.56 &                      10.9\% \\
 \rowcolor{lightgreen}                Gothic-PROIEL &               big &          92.39 &      96.10 &            96.21 &                      50.2\% \\
 \rowcolor{lightgreen}                    Greek-GDT &               big &          94.74 &      97.22 &            97.26 &                      47.9\% \\
 \rowcolor{lightgreen}                   Hebrew-HTB &               big &          82.88 &      82.90 &            82.93 &                       0.3\% \\
 \rowcolor{lightgreen}                   Hindi-HDTB &               big &          98.45 &      98.68 &            98.70 &                      16.1\% \\
 \rowcolor{lightgreen}             Hungarian-Szeged &               big &          92.99 &      94.53 &            94.57 &                      22.5\% \\
 \rowcolor{lightgreen}               Indonesian-GSD &               big &          99.60 &      99.69 &            99.68 &                      20.0\% \\
 \rowcolor{lightgreen}                    Irish-IDT &             small &          87.52 &      90.62 &            90.52 &                      24.0\% \\
                                       Italian-ISDT &               big &          98.21 &      98.09 &            98.16 &                      -2.7\% \\
 \rowcolor{lightgreen}             Italian-PoSTWITA &               big &          94.91 &      96.61 &            96.63 &                      33.8\% \\
                                       Japanese-GSD &               big &          90.01 &      89.94 &            89.63 &                      -3.7\% \\
 \rowcolor{lightgreen}                   Kazakh-KTB &               low &          57.36 &      48.61 &            57.43 &                       0.2\% \\
 \rowcolor{lightgreen}                   Korean-GSD &               big &          91.37 &      93.83 &            93.94 &                      29.8\% \\
 \rowcolor{lightgreen}                 Korean-Kaist &               big &          93.53 &      94.38 &            94.39 &                      13.3\% \\
 \rowcolor{lightgreen}                  Kurmanji-MG &               low &          52.44 &      42.45 &            64.83 &                      26.1\% \\
 \rowcolor{lightgreen}                   Latin-ITTB &               big &          98.56 &      98.66 &            98.67 &                       7.6\% \\
 \rowcolor{lightgreen}                 Latin-PROIEL &               big &          95.54 &      97.14 &            97.20 &                      37.2\% \\
 \rowcolor{lightgreen}                Latin-Perseus &             small &          75.44 &      85.37 &            85.27 &                      40.0\% \\
 \rowcolor{lightgreen}                 Latvian-LVTB &               big &          93.33 &      93.69 &            93.95 &                       9.3\% \\
 \rowcolor{lightgreen}            North Sami-Giella &             small &          78.43 &      89.54 &            89.70 &                      52.2\% \\
                                  Norwegian-Bokmaal &               big &          98.20 &      97.87 &            97.97 &                     -11.3\% \\
                                  Norwegian-Nynorsk &               big &          97.80 &      97.71 &            97.72 &                      -3.5\% \\
 \rowcolor{lightgreen}         Norwegian-NynorskLIA &             small &          92.65 &      92.91 &            94.51 &                      25.3\% \\
 \rowcolor{lightgreen}   Old Church Slavonic-PROIEL &               big &          88.93 &      95.33 &            95.14 &                      56.1\% \\
                                     Persian-Seraji &               big &          97.05 &      96.99 &            96.77 &                      -8.7\% \\
 \rowcolor{lightgreen}                   Polish-LFG &               big &          96.73 &      97.50 &            97.66 &                      28.4\% \\
 \rowcolor{lightgreen}                    Polish-SZ &               big &          95.31 &      96.93 &            97.08 &                      37.7\% \\
 \rowcolor{lightgreen}            Portuguese-Bosque &               big &          97.38 &      97.53 &            97.58 &                       7.6\% \\
 \rowcolor{lightgreen}                 Romanian-RRT &               big &          97.61 &      98.25 &            98.23 &                      25.9\% \\
 \rowcolor{lightgreen}            Russian-SynTagRus &               big &          97.94 &      98.16 &            98.15 &                      10.2\% \\
 \rowcolor{lightgreen}                Russian-Taiga &             small &          83.55 &      88.47 &            89.32 &                      35.1\% \\
 \rowcolor{lightgreen}                  Serbian-SET &               big &          96.56 &      97.09 &            97.17 &                      17.7\% \\
 \rowcolor{lightgreen}                   Slovak-SNK &               big &          95.66 &      96.27 &            96.35 &                      15.9\% \\
 \rowcolor{lightgreen}                Slovenian-SSJ &               big &          96.22 &      96.35 &            96.49 &                       7.1\% \\
 \rowcolor{lightgreen}                Slovenian-SST &             small &          92.56 &      95.06 &            94.90 &                      31.5\% \\
                                     Spanish-AnCora &               big &          99.02 &      98.45 &            98.48 &                     -35.5\% \\
 \rowcolor{lightgreen}                Swedish-LinES &               big &          96.61 &      96.87 &            97.29 &                      20.1\% \\
 \rowcolor{lightgreen}                  Swedish-PUD &               PUD &          86.23 &      86.69 &            87.47 &                       9.0\% \\
 \rowcolor{lightgreen}            Swedish-Talbanken &               big &          97.08 &      97.81 &            97.98 &                      30.8\% \\
 \rowcolor{lightgreen}                 Turkish-IMST &               big &          92.74 &      94.85 &            95.16 &                      33.3\% \\
 \rowcolor{lightgreen}                 Ukrainian-IU &               big &          95.94 &      96.52 &            96.62 &                      16.7\% \\
                                 Upper Sorbian-UFAL &               low &          63.54 &      53.73 &            54.80 &                     -19.3\% \\
 \rowcolor{lightgreen}                    Urdu-UDTB &               big &          97.33 &      97.42 &            97.43 &                       3.7\% \\
 \rowcolor{lightgreen}                   Uyghur-UDT &               big &          92.86 &      94.09 &            94.15 &                      18.1\% \\
                                     Vietnamese-VTB &               big &          84.76 &      84.16 &            84.26 &                      -3.2\% \\\hline
 \rowcolor{lightgreen}                \textbf{Average} &                   &          91.64 &      92.22 &            93.24 &                         19.1\% \\
\hline\hline
%\end{tabular}
\\
\caption{Lemmatization accuracies for all 76 treebanks studied in this paper measured on test data with predicted segmentation. Green color indicates treebanks where our overall best method, Augm. Mixed 8K + 8K, outperforms the best overall baseline, UDPipe Future.}
\label{tbl:big_table}
\end{longtable}
}}
%%% MACHINE GENERATED TABLE %%%

While the autoencoding augmentation method does not require any additional data, the transducer-based techniques move the system into an unconstrained setting, if considering a task setup where only the given treebanks are allowed in system training. However, in real-life situations, where all available data is allowed, the comparison between our augmented system and the baseline systems is fair. Such a real-life task setting was used for example in the CoNLL 2018 and 2017 multilingual parsing shared tasks, where a list of additional resources apart from the treebanks were given to all task participants. These allowed resources also included the Apertium morphological transducers, which makes the comparison between our augmentation methods and baseline systems from the CoNLL 2018 shared task fair. The difference between our system and a standard context-based lemmatization system is that integrating information from these additional sources is much easier with our task setting where the lemmatizer does not need the words to appear in a natural context.

\subsection{Generalization and Error Propagation}

To understand the generalization capability of the lemmatizer when the segmentation and morphological tagging effects are disregarded, we compare the lemmatization accuracy on top of predicted segmentation to gold-standard segmentation (sentence and word level), as well as on top of predicted morphosyntactic features to gold-standard morphosyntactic features. The same experiment also measures the risk of error propagation, where the lemmatizer makes a mistake due to incorrectly predicted morphosyntactic features. Results for all treebanks are available in Appendix~\ref{app:eval}. When comparing the lemmatization accuracy of the 5 low-resource languages (Armenian, Buryat, Kazakh, Kurmanji, Upper Sorbian) on predicted and gold morphosyntactic features, the four transducer languages (Armenian, Buryat, Kazakh, Kurmanji) appear to generalize extremely well, gold morphosyntactic features increasing the accuracy from 58\%---74\% to 91\%---96\%. For Upper Sorbian, the one low-resource language without a transducer, the generalization ability is clearly worse, gold tags increasing accuracy only from 55\% to 74\%. These results suggest that the data augmentation techniques utilizing a morphological transducer are sufficient enough to train a high quality lemmatizer if reliable morphosyntactic features are available. However, at the same time it shows that in extreme cases where the accuracy of part-of-speech tagging is barely above 50\%, errors from the tagger component propagate notably. As a future work, it would be interesting to study whether morphological transducers could be used to create artificial data for context-dependent morphological tagging so as to improve the tagger performance as well.

% Armenian: 73.77 --> 93.47 (dictionary baseline (form): 61.89, transducer coverage: 64.61, transducer recall: 61.82)
% Buryat: 58.11 --> 90.99 (dictionary baseline (form): 56.43, transducer coverage: 99.06, transducer recall: 97.11)
% Kazakh: 61.83 --> 95.66 (dictionary baseline (form): 53.00, transducer coverage: 98.15, transducer recall: 97.35)
% Kurmanji: 67.94 --> 91.72 (dictionary baseline (form): 59.51, transducer coverage: 4.70, transducer recall: 4.70) --- here train data is 20 sentences
% Upper Sorbian: 55.26 --> 73.74 (dictionary baseline (form): 60.23, no transducer)

Currently, the lemmatizer is the last component in the parsing pipeline, thus not affecting the labeled attachment score of the syntactic parser. The parser currently used in the pipeline was originally designed to not consider lemmas at all, however, the lemmatizer component could be located before the syntactic parser as well, making it possible to establish whether using lemmas as additional features during parsing would improve its performance.

\subsection{Future Work}

We acknowledge that the morphological transducers used in our data augmentation study may not have been utilized to their full power. Our straightforward feature mapping from the Apertium framework into Universal Dependencies was designed to be language agnostic, thus suffering from inconsistencies in annotations between different languages and treebanks. A more focused attempt on a particular, well chosen language with an improved morphological transducer, language specific conversion or detailed parameter tuning could yield better results. While Apertium can be considered a trustworthy source for unified morphological resources, for many languages, more developed language-specific transducers exist. For example, if particularly working on Turkish, Finnish or Hungarian, one should consider using morphological transducers by \citet{coltekin2010freely}, \citet{pirinen2015development} and \citet{tron2006morphdb}. A focused per-language effort is naturally entirely out-of-scope of this current work, which can nevertheless serve as a basis for such a language-specific development. Similar argumentation is suggested by \citet{pirinen2019neural}, who carried out a focused evaluation of our lemmatization system and the OMorFi morphological analyzer \citep{pirinen2015development} on the Finnish language. OMorFi is a mature system, being the result of a major development effort spanning over several years. Its output is in the Universal Dependencies scheme, providing a valid point of comparison. A lemmatization performance of our pipeline far superior to that of OMorFi is reported, leading to the conclusion that the machine learning approach is indeed highly competitive with the traditional transducers and can be seen as the preferred approach to developing lemmatizers for new languages. However, we leave it as a future work to study whether combining such a morphological transducer and machine learning approach in a targeted data augmentation effort would yield higher improvements for lemmatization accuracy than presented in this paper.

Another interesting direction to expand the work in future would be to test how well the lemmatizer works on short text segments, for example with search queries, where deep learning systems traditionally need to be trained separately to match the different style of writing, for example very often omitting the main verb. As the lemmatizer is operating on the word level without a notion of context, this should not pose an issue during the lemmatization. However, a separate question is how reliable a morphological tagger would be with such short text segments.

\subsection{Model and Software Release}

We release trained models for all 76 treebanks experimented in this paper, embedded into a full parsing pipeline including segmentation, tagging, syntactic parsing and lemmatization. The parsing pipeline source code is available at \mbox{\url{https://turkunlp.org/Turku-neural-parser-pipeline}} under the Apache 2.0 license. It includes trained models for all the necessary components (segmentation, tagging, syntactic parsing and lemmatization), trained on the UD v2.2 treebanks. The whole processing pipeline can be executed with a single command, removing the need for data reformatting between the different analysis components. The pipeline runs in a Python environment which can be installed with or without GPU support. To increase the usability across different platforms we also provide a publicly accessible Docker image, which wraps the pipeline in a container which can be executed without manual installation, assuring that the pipeline can be executed and the results replicated also in the future.

\subsection{Training and Prediction Speed}
\label{sec:speed}

Typical training times for the lemmatizer models on UD treebanks with 50 training epochs are 1-2 hours on one Nvidia GeForce K80 GPU card. The largest treebanks (Czech-PDT 1.2M tokens and Russian-SynTagRus 870K tokens) took approximately 15 hours to train for the full 50 epochs. However, the training usually converges between epochs 30 and 40, and therefore training time could be reduced using an early stopping criterion.

In prediction time, we present several advantages over previous sequence-to-sequence lemmatizer models. First, by using morphosyntactic features instead of a sliding window of text to represent the contextual information, after running the context-dependent morphological tagger, the lemmatizer is able to process each word independently from its textual sentence context, and therefore we only need to lemmatize each unique word and feature combination. This enables us to 1) only lemmatize unique items inside each textual batch, and 2) store a cache of common pre-analysed words, and only run the sequence-to-sequence model for words not already present in this global lemma cache. Together with the trained models we distribute such a global cache file for each language.

Prediction times for the full parsing pipeline, including segmentation, tagging, syntactic parsing and lemmatization, are on the order of 1,300 tokens per second (about 100 sentences per second) on an Nvidia GeForce GTX 1070 card. On a server-grade CPU-only computer (24 cores and 250GB RAM) prediction times are 350 tokens per second, while on a consumer CPU-only laptop (8 cores and 8GB of RAM), the full pipeline can process about 280 tokens per second. These are measured with a pre-analysed lemma cache collected from the training data, and prediction times especially on CPU could be yet improved by collecting a larger pre-analysed lemma cache using, for example, large web corpora.

\section{Conclusions}
\label{sec:conclusions}

In this paper we have introduced a novel sequence-to-sequence lemmatization method utilizing morphosyntactic tags to inform the model about the context of the word. We validated the hypothesis that the tags provide a sufficient disambiguation context using statistics from the Universal Dependencies treebanks across a large number of languages. We presented a careful evaluation of our method over several baselines and 52 different languages showing that the method surpasses all the baseline systems, reducing relative errors on average by 19\% across 76 treebanks compared to the best overall baseline. The lemmatizer presented in this work was also used as our entry in the CoNLL-18 Shared Task on Multilingual Parsing from Raw Text to Universal Dependencies, where we achieved the 1st place out of 26 teams on two evaluation metrics incorporating lemmatization. Additionally, we investigated two different data augmentation methods to boost the lemmatization performance of our base system. We found that augmenting the training data using a mixture of autoencoder training and the output of a morphological transducer decreases the error rate by 13\% relative to the un-augmented system, with the gain being unsurprisingly concentrated on the low-resource languages.

As an overall conclusion, we have demonstrated a highly competitive performance of the generic sequence-to-sequence paradigm on the lemmatization task, surpassing in accuracy prior methods specifically developed for lemmatization. 

The lemmatization models for all languages reported in the paper, source code and materials for all experiments, the full parsing pipeline source code and parsing models, as well as an easy-to-use Docker container, are available at \url{https://turkunlp.org/Turku-neural-parser-pipeline} under the Apache 2.0 license.

\section*{Acknowledgments}

We gratefully acknowledge the support of Academy of Finland, CSC – IT Center for Science, and the NVIDIA Corporation GPU Grant Program.

\bibliography{main}

\begin{thebibliography}{}
\expandafter\ifx\csname natexlab\endcsname\relax\def\natexlab#1{#1}\fi

\bibitem[{Aker et~al.(2017)Aker, Petrak, and Sabbah}]{aker2017extensible}
Ahmet Aker, Johann Petrak, and Firas Sabbah. 2017.
\newblock An extensible multilingual open source lemmatizer.
\newblock In {\em Proceedings of the International Conference Recent Advances
  in Natural Language Processing, RANLP 2017\/}. INCOMA Ltd., Varna, Bulgaria,
  pages 40--45.

\bibitem[{Bergmanis and Goldwater(2018)}]{bergmanis2018context}
Toms Bergmanis and Sharon Goldwater. 2018.
\newblock Context sensitive neural lemmatization with {L}ematus.
\newblock In {\em Proceedings of the 2018 Conference of the North American
  Chapter of the Association for Computational Linguistics: Human Language
  Technologies, Volume 1 (Long Papers)\/}. Association for Computational
  Linguistics, New Orleans, Louisiana, volume~1, pages 1391--1400.

\bibitem[{Bergmanis et~al.(2017)Bergmanis, Kann, Sch{\"u}tze, and
  Goldwater}]{bergmanis2017training}
Toms Bergmanis, Katharina Kann, Hinrich Sch{\"u}tze, and Sharon Goldwater.
  2017.
\newblock Training data augmentation for low-resource morphological inflection.
\newblock {\em Proceedings of the CoNLL SIGMORPHON 2017 Shared Task: Universal
  Morphological Reinflection\/} pages 31--39.

\bibitem[{Chakrabarty et~al.(2017)Chakrabarty, Pandit, and
  Garain}]{chakrabarty2017context}
Abhisek Chakrabarty, Onkar~Arun Pandit, and Utpal Garain. 2017.
\newblock Context sensitive lemmatization using two successive bidirectional
  gated recurrent networks.
\newblock In {\em Proceedings of the 55th Annual Meeting of the Association for
  Computational Linguistics (Volume 1: Long Papers)\/}. Association for
  Computational Linguistics, Vancouver, Canada, volume~1, pages 1481--1491.

\bibitem[{Chrupa{\l}a et~al.(2008)Chrupa{\l}a, Dinu, and
  Van~Genabith}]{chrupala2008learning}
Grzegorz Chrupa{\l}a, Georgiana Dinu, and Josef Van~Genabith. 2008.
\newblock Learning morphology with {M}orfette.
\newblock In {\em Proceedings of the Sixth International Conference on Language
  Resources and Evaluation (LREC'08)\/}. European Language Resources
  Association (ELRA), Marrakech, Morocco, pages 2362--2367.

\bibitem[{{\c{C}}{\"o}ltekin(2010)}]{coltekin2010freely}
{\c{C}}a{\u{g}}r{\i} {\c{C}}{\"o}ltekin. 2010.
\newblock A freely available morphological analyzer for {T}urkish.
\newblock In {\em Proceedings of the Seventh International Conference on
  Language Resources and Evaluation (LREC'10)\/}. European Language Resources
  Association (ELRA), Valletta, Malta, volume~2, pages 19--28.

\bibitem[{Costa and Nannicini(2018)}]{costa2014rbfopt}
Alberto Costa and Giacomo Nannicini. 2018.
\newblock {RBFOpt}: an open-source library for black-box optimization with
  costly function evaluations.
\newblock {\em Mathematical Programming Computation\/} 10:597--629.

\bibitem[{Cotterell et~al.(2017)Cotterell, Kirov, Sylak-Glassman, Walther,
  Vylomova, Xia, Faruqui, K\"ubler, Yarowsky, Eisner, and
  Hulden}]{cotterell-EtAl:2017:K17-20}
Ryan Cotterell, Christo Kirov, John Sylak-Glassman, G\'eraldine Walther,
  Ekaterina Vylomova, Patrick Xia, Manaal Faruqui, Sandra K\"ubler, David
  Yarowsky, Jason Eisner, and Mans Hulden. 2017.
\newblock {CoNLL-SIGMORPHON 2017 Shared Task: Universal Morphological
  Reinflection in 52 Languages}.
\newblock In {\em Proceedings of the CoNLL SIGMORPHON 2017 Shared Task:
  Universal Morphological Reinflection\/}. Association for Computational
  Linguistics, Vancouver, Canada, pages 1--30.

\bibitem[{Dozat and Manning(2017)}]{dozat2017deep}
Timothy Dozat and Christopher~D Manning. 2017.
\newblock Deep biaffine attention for neural dependency parsing.
\newblock In {\em Proceedings of the 2017 International Conference on Learning
  Representations (ICLR'17)\/}.

\bibitem[{Dozat et~al.(2017)Dozat, Qi, and Manning}]{dozat2017stanford}
Timothy Dozat, Peng Qi, and Christopher~D Manning. 2017.
\newblock Stanford's graph-based neural dependency parser at the {CoNLL 2017
  Shared Task}.
\newblock In {\em Proceedings of the CoNLL 2017 Shared Task: Multilingual
  Parsing from Raw Text to Universal Dependencies\/}. Association for
  Computational Linguistics, pages 20--30.

\bibitem[{Ginter et~al.(2017)Ginter, Haji{\v c}, Luotolahti, Straka, and
  Zeman}]{conll-raw-data}
Filip Ginter, Jan Haji{\v c}, Juhani Luotolahti, Milan Straka, and Daniel
  Zeman. 2017.
\newblock {CoNLL 2017 Shared Task} - automatically annotated raw texts and word
  embeddings.
\newblock {LINDAT}/{CLARIN} digital library at the {Institute of Formal and
  Applied Linguistics} ({{\'U}FAL}), {Faculty of Mathematics and Physics,
  Charles University}.

\bibitem[{Jean et~al.(2015)Jean, Cho, Memisevic, and Bengio}]{jean2015using}
S{\'e}bastien Jean, Kyunghyun Cho, Roland Memisevic, and Yoshua Bengio. 2015.
\newblock On using very large target vocabulary for neural machine translation.
\newblock In {\em Proceedings of the 53rd Annual Meeting of the Association for
  Computational Linguistics (ACL 2015)\/}. Association for Computational
  Linguistics, Beijing, China, pages 1--10.

\bibitem[{Kanerva et~al.(2018)Kanerva, Ginter, Miekka, Leino, and
  Salakoski}]{udst:turkunlp}
Jenna Kanerva, Filip Ginter, Niko Miekka, Akseli Leino, and Tapio Salakoski.
  2018.
\newblock Turku neural parser pipeline: An end-to-end system for the {CoNLL}
  2018 {S}hared {T}ask.
\newblock In {\em Proceedings of the CoNLL 2018 Shared Task: Multilingual
  Parsing from Raw Text to Universal Dependencies\/}. Association for
  Computational Linguistics, Brussels, Belgium.

\bibitem[{Kann and Sch{\"u}tze(2017{\natexlab{a}})}]{kann2017lmu}
Katharina Kann and Hinrich Sch{\"u}tze. 2017{\natexlab{a}}.
\newblock The {LMU} system for the {CoNLL-SIGMORPHON 2017 Shared Task on
  Universal Morphological Reinflection}.
\newblock In {\em Proceedings of the CoNLL SIGMORPHON 2017 Shared Task:
  Universal Morphological Reinflection\/}. Association for Computational
  Linguistics, Vancouver, Canada, pages 40--48.

\bibitem[{Kann and Sch{\"u}tze(2017{\natexlab{b}})}]{kann2017unlabeled}
Katharina Kann and Hinrich Sch{\"u}tze. 2017{\natexlab{b}}.
\newblock Unlabeled data for morphological generation with character-based
  sequence-to-sequence models.
\newblock In {\em Proceedings of the 1st Workshop on Subword and Character
  Level Models in NLP (SCLeM 2017)\/}. Association for Computational
  Linguistics, Copenhagen, Denmark.

\bibitem[{Karttunen and Beesley(1992)}]{karttunen1992two}
Lauri Karttunen and Kenneth~R Beesley. 1992.
\newblock {\em Two-level rule compiler\/}.
\newblock Xerox Corporation. Palo Alto Research Center.

\bibitem[{Kingma and Ba(2015)}]{kingma2015adam}
Diederik Kingma and Jimmy Ba. 2015.
\newblock Adam: A method for stochastic optimization.
\newblock In {\em Proceedings of the 3rd International Conference for Learning
  Representations\/}.

\bibitem[{Kirov et~al.(2016)Kirov, Sylak-Glassman, Que, and
  Yarowsky}]{unimorph2016}
Christo Kirov, John Sylak-Glassman, Roger Que, and David Yarowsky. 2016.
\newblock Very-large scale parsing and normalization of {W}iktionary
  morphological paradigms.
\newblock In Nicoletta Calzolari~(Conference Chair), Khalid Choukri, Thierry
  Declerck, Sara Goggi, Marko Grobelnik, Bente Maegaard, Joseph Mariani, Helene
  Mazo, Asuncion Moreno, Jan Odijk, and Stelios Piperidis, editors, {\em
  Proceedings of the Tenth International Conference on Language Resources and
  Evaluation (LREC 2016)\/}. European Language Resources Association (ELRA),
  Paris, France.

\bibitem[{Klein et~al.(2017)Klein, Kim, Deng, Senellart, and Rush}]{opennmt}
Guillaume Klein, Yoon Kim, Yuntian Deng, Jean Senellart, and Alexander~M. Rush.
  2017.
\newblock {OpenNMT}: Open-source toolkit for neural machine translation.
\newblock In {\em Proceedings of the 55th annual meeting of the Association for
  Computational Linguistics (ACL'17)\/}. Association for Computational
  Linguistics, Vancouver, Canada.

\bibitem[{Kondratyuk et~al.(2018)Kondratyuk, Gaven{\v{c}}iak, Straka, and
  Haji{\v{c}}}]{kondratyuk2018lemmatag}
Daniel Kondratyuk, Tom{\'a}{\v{s}} Gaven{\v{c}}iak, Milan Straka, and Jan
  Haji{\v{c}}. 2018.
\newblock {LemmaTag}: Jointly tagging and lemmatizing for morphologically rich
  languages with {BRNNs}.
\newblock In {\em Proceedings of the 2018 Conference on Empirical Methods in
  Natural Language Processing\/}. Association for Computational Linguistics,
  pages 4921--4928.

\bibitem[{Koskenniemi(1984)}]{koskenniemi1984general}
Kimmo Koskenniemi. 1984.
\newblock A general computational model for word-form recognition and
  production.
\newblock In {\em Proceedings of the 10th international conference on
  Computational Linguistics\/}. Association for Computational Linguistics, USA,
  pages 178--181.

\bibitem[{Liu and Hulden(2017)}]{liu2017evaluation}
Ling Liu and Mans Hulden. 2017.
\newblock Evaluation of finite state morphological analyzers based on paradigm
  extraction from {W}iktionary.
\newblock In {\em Proceedings of the 13th International Conference on Finite
  State Methods and Natural Language Processing (FSMNLP 2017)\/}. Association
  for Computational Linguistics, Ume{\aa}, Sweden, pages 69--74.

\bibitem[{Luong et~al.(2015{\natexlab{a}})Luong, Sutskever, Le, Vinyals, and
  Zaremba}]{luong2015addressing}
Minh-Thang Luong, Ilya Sutskever, Quoc~V Le, Oriol Vinyals, and Wojciech
  Zaremba. 2015{\natexlab{a}}.
\newblock Addressing the rare word problem in neural machine translation.
\newblock In {\em Proceedings of the 53rd Annual Meeting of the Association for
  Computational Linguistics (ACL 2015)\/}. Association for Computational
  Linguistics, Beijing, China, pages 11--19.

\bibitem[{Luong et~al.(2015{\natexlab{b}})Luong, Pham, and
  Manning}]{luong2015EMNLP}
Thang Luong, Hieu Pham, and Christopher~D. Manning. 2015{\natexlab{b}}.
\newblock Effective approaches to attention-based neural machine translation.
\newblock In {\em Proceedings of the 2015 Conference on Empirical Methods in
  Natural Language Processing\/}. Association for Computational Linguistics,
  Lisbon, Portugal, pages 1412--1421.

\bibitem[{McCarthy et~al.(2018)McCarthy, Silfverberg, Cotterell, Hulden, and
  Yarowsky}]{mccarthy2018marrying}
Arya~D. McCarthy, Miikka Silfverberg, Ryan Cotterell, Mans Hulden, and David
  Yarowsky. 2018.
\newblock Marrying {U}niversal {D}ependencies and {U}niversal {M}orphology.
\newblock In {\em Proceedings of the 2018 Workshop on Universal Dependencies
  (UDW 2018)\/}. Association for Computational Linguistics.

\bibitem[{M{\"u}ller et~al.(2015)M{\"u}ller, Cotterell, Fraser, and
  Sch{\"u}tze}]{muller2015joint}
Thomas M{\"u}ller, Ryan Cotterell, Alexander Fraser, and Hinrich Sch{\"u}tze.
  2015.
\newblock Joint lemmatization and morphological tagging with {L}emming.
\newblock In {\em Proceedings of the 2015 Conference on Empirical Methods in
  Natural Language Processing\/}. Association for Computational Linguistics,
  Lisbon, Portugal, pages 2268--2274.

\bibitem[{Nivre et~al.(2018)Nivre, Abrams, Agi{\'c}, Ahrenberg, Antonsen,
  Aranzabe, Arutie, Asahara, Ateyah, Attia, Atutxa, Augustinus, Badmaeva,
  Ballesteros, Banerjee, Bank, Barbu~Mititelu, Bauer, Bellato, Bengoetxea,
  Bhat, Biagetti, Bick, Blokland, Bobicev, B{\"o}rstell, Bosco, Bouma, Bowman,
  Boyd, Burchardt, Candito, Caron, Caron, Cebiro{\u g}lu~Eryi{\u g}it, Celano,
  Cetin, Chalub, Choi, Cho, Chun, Cinkov{\'a}, Collomb, {\c C}{\"o}ltekin,
  Connor, Courtin, Davidson, de~Marneffe, de~Paiva, Diaz~de Ilarraza,
  Dickerson, Dirix, Dobrovoljc, Dozat, Droganova, Dwivedi, Eli, Elkahky,
  Ephrem, Erjavec, Etienne, Farkas, Fernandez~Alcalde, Foster, Freitas,
  Gajdo{\v s}ov{\'a}, Galbraith, Garcia, G{\"a}rdenfors, Gerdes, Ginter,
  Goenaga, Gojenola, G{\"o}k{\i}rmak, Goldberg, G{\'o}mez~Guinovart,
  Gonz{\'a}les~Saavedra, Grioni, Gr{\=u}z{\={\i}}tis, Guillaume,
  Guillot-Barbance, Habash, Haji{\v c}, Haji{\v c}~jr., H{\`a}~M{\~y}, Han,
  Harris, Haug, Hladk{\'a}, Hlav{\'a}{\v c}ov{\'a}, Hociung, Hohle, Hwang, Ion,
  Irimia, Jel{\'{\i}}nek, Johannsen, J{\o}rgensen, Ka{\c s}{\i}kara, Kahane,
  Kanayama, Kanerva, Kayadelen, Kettnerov{\'a}, Kirchner, Kotsyba, Krek, Kwak,
  Laippala, Lambertino, Lando, Larasati, Lavrentiev, Lee, L{\^e}~H{\`{\^o}}ng,
  Lenci, Lertpradit, Leung, Li, Li, Li, Lim, Ljube{\v s}i{\'c}, Loginova,
  Lyashevskaya, Lynn, Macketanz, Makazhanov, Mandl, Manning, Manurung, M{\u
  a}r{\u a}nduc, Mare{\v c}ek, Marheinecke, Mart{\'{\i}}nez~Alonso, Martins,
  Ma{\v s}ek, Matsumoto, {McDonald}, Mendon{\c c}a, Miekka, Missil{\"a},
  Mititelu, Miyao, Montemagni, More, Moreno~Romero, Mori, Mortensen,
  Moskalevskyi, Muischnek, Murawaki, M{\"u}{\"u}risep, Nainwani,
  Navarro~Hor{\~n}iacek, Nedoluzhko, Ne{\v s}pore-B{\=e}rzkalne,
  Nguy{\~{\^e}}n~Th{\d i}, Nguy{\~{\^e}}n Th{\d i}~Minh, Nikolaev, Nitisaroj,
  Nurmi, Ojala, Ol{\'u}{\`o}kun, Omura, Osenova, {\"O}stling, {\O}vrelid,
  Partanen, Pascual, Passarotti, Patejuk, Peng, Perez, Perrier, Petrov,
  Piitulainen, Pitler, Plank, Poibeau, Popel, Pretkalni{\c n}a, Pr{\'e}vost,
  Prokopidis, Przepi{\'o}rkowski, Puolakainen, Pyysalo, R{\"a}{\"a}bis,
  Rademaker, Ramasamy, Rama, Ramisch, Ravishankar, Real, Reddy, Rehm,
  Rie{\ss}ler, Rinaldi, Rituma, Rocha, Romanenko, Rosa, Rovati, Roșca, Rudina,
  Sadde, Saleh, Samard{\v z}i{\'c}, Samson, Sanguinetti, Saul{\={\i}}te,
  Sawanakunanon, Schneider, Schuster, Seddah, Seeker, Seraji, Shen, Shimada,
  Shohibussirri, Sichinava, Silveira, Simi, Simionescu, Simk{\'o}, {\v
  S}imkov{\'a}, Simov, Smith, Soares-Bastos, Stella, Straka, Strnadov{\'a},
  Suhr, Sulubacak, Sz{\'a}nt{\'o}, Taji, Takahashi, Tanaka, Tellier, Trosterud,
  Trukhina, Tsarfaty, Tyers, Uematsu, Ure{\v s}ov{\'a}, Uria, Uszkoreit,
  Vajjala, van Niekerk, van Noord, Varga, Vincze, Wallin, Washington, Williams,
  Wir{\'e}n, Woldemariam, Wong, Yan, Yavrumyan, Yu, {\v Z}abokrtsk{\'y},
  Zeldes, Zeman, Zhang, and Zhu}]{udv2.2}
Joakim Nivre, Mitchell Abrams, {\v Z}eljko Agi{\'c}, Lars Ahrenberg, Lene
  Antonsen, Maria~Jesus Aranzabe, Gashaw Arutie, Masayuki Asahara, Luma Ateyah,
  Mohammed Attia, Aitziber Atutxa, Liesbeth Augustinus, Elena Badmaeva, Miguel
  Ballesteros, Esha Banerjee, Sebastian Bank, Verginica Barbu~Mititelu, John
  Bauer, Sandra Bellato, Kepa Bengoetxea, Riyaz~Ahmad Bhat, Erica Biagetti,
  Eckhard Bick, Rogier Blokland, Victoria Bobicev, Carl B{\"o}rstell, Cristina
  Bosco, Gosse Bouma, Sam Bowman, Adriane Boyd, Aljoscha Burchardt, Marie
  Candito, Bernard Caron, Gauthier Caron, G{\"u}l{\c s}en Cebiro{\u
  g}lu~Eryi{\u g}it, Giuseppe G.~A. Celano, Savas Cetin, Fabricio Chalub, Jinho
  Choi, Yongseok Cho, Jayeol Chun, Silvie Cinkov{\'a}, Aur{\'e}lie Collomb, {\c
  C}a{\u g}r{\i} {\c C}{\"o}ltekin, Miriam Connor, Marine Courtin, Elizabeth
  Davidson, Marie-Catherine de~Marneffe, Valeria de~Paiva, Arantza Diaz~de
  Ilarraza, Carly Dickerson, Peter Dirix, Kaja Dobrovoljc, Timothy Dozat, Kira
  Droganova, Puneet Dwivedi, Marhaba Eli, Ali Elkahky, Binyam Ephrem, Toma{\v
  z} Erjavec, Aline Etienne, Rich{\'a}rd Farkas, Hector Fernandez~Alcalde,
  Jennifer Foster, Cl{\'a}udia Freitas, Katar{\'{\i}}na Gajdo{\v s}ov{\'a},
  Daniel Galbraith, Marcos Garcia, Moa G{\"a}rdenfors, Kim Gerdes, Filip
  Ginter, Iakes Goenaga, Koldo Gojenola, Memduh G{\"o}k{\i}rmak, Yoav Goldberg,
  Xavier G{\'o}mez~Guinovart, Berta Gonz{\'a}les~Saavedra, Matias Grioni,
  Normunds Gr{\=u}z{\={\i}}tis, Bruno Guillaume, C{\'e}line Guillot-Barbance,
  Nizar Habash, Jan Haji{\v c}, Jan Haji{\v c}~jr., Linh H{\`a}~M{\~y}, Na-Rae
  Han, Kim Harris, Dag Haug, Barbora Hladk{\'a}, Jaroslava Hlav{\'a}{\v
  c}ov{\'a}, Florinel Hociung, Petter Hohle, Jena Hwang, Radu Ion, Elena
  Irimia, Tom{\'a}{\v s} Jel{\'{\i}}nek, Anders Johannsen, Fredrik
  J{\o}rgensen, H{\"u}ner Ka{\c s}{\i}kara, Sylvain Kahane, Hiroshi Kanayama,
  Jenna Kanerva, Tolga Kayadelen, V{\'a}clava Kettnerov{\'a}, Jesse Kirchner,
  Natalia Kotsyba, Simon Krek, Sookyoung Kwak, Veronika Laippala, Lorenzo
  Lambertino, Tatiana Lando, Septina~Dian Larasati, Alexei Lavrentiev, John
  Lee, Phuong L{\^e}~H{\`{\^o}}ng, Alessandro Lenci, Saran Lertpradit, Herman
  Leung, Cheuk~Ying Li, Josie Li, Keying Li, {KyungTae} Lim, Nikola Ljube{\v
  s}i{\'c}, Olga Loginova, Olga Lyashevskaya, Teresa Lynn, Vivien Macketanz,
  Aibek Makazhanov, Michael Mandl, Christopher Manning, Ruli Manurung, C{\u
  a}t{\u a}lina M{\u a}r{\u a}nduc, David Mare{\v c}ek, Katrin Marheinecke,
  H{\'e}ctor Mart{\'{\i}}nez~Alonso, Andr{\'e} Martins, Jan Ma{\v s}ek, Yuji
  Matsumoto, Ryan {McDonald}, Gustavo Mendon{\c c}a, Niko Miekka, Anna
  Missil{\"a}, C{\u a}t{\u a}lin Mititelu, Yusuke Miyao, Simonetta Montemagni,
  Amir More, Laura Moreno~Romero, Shinsuke Mori, Bjartur Mortensen, Bohdan
  Moskalevskyi, Kadri Muischnek, Yugo Murawaki, Kaili M{\"u}{\"u}risep, Pinkey
  Nainwani, Juan~Ignacio Navarro~Hor{\~n}iacek, Anna Nedoluzhko, Gunta Ne{\v
  s}pore-B{\=e}rzkalne, Luong Nguy{\~{\^e}}n~Th{\d i}, Huy{\`{\^e}}n
  Nguy{\~{\^e}}n Th{\d i}~Minh, Vitaly Nikolaev, Rattima Nitisaroj, Hanna
  Nurmi, Stina Ojala, Ad{\'e}day{\d o} Ol{\'u}{\`o}kun, Mai Omura, Petya
  Osenova, Robert {\"O}stling, Lilja {\O}vrelid, Niko Partanen, Elena Pascual,
  Marco Passarotti, Agnieszka Patejuk, Siyao Peng, Cenel-Augusto Perez, Guy
  Perrier, Slav Petrov, Jussi Piitulainen, Emily Pitler, Barbara Plank, Thierry
  Poibeau, Martin Popel, Lauma Pretkalni{\c n}a, Sophie Pr{\'e}vost, Prokopis
  Prokopidis, Adam Przepi{\'o}rkowski, Tiina Puolakainen, Sampo Pyysalo,
  Andriela R{\"a}{\"a}bis, Alexandre Rademaker, Loganathan Ramasamy, Taraka
  Rama, Carlos Ramisch, Vinit Ravishankar, Livy Real, Siva Reddy, Georg Rehm,
  Michael Rie{\ss}ler, Larissa Rinaldi, Laura Rituma, Luisa Rocha, Mykhailo
  Romanenko, Rudolf Rosa, Davide Rovati, Valentin Roșca, Olga Rudina, Shoval
  Sadde, Shadi Saleh, Tanja Samard{\v z}i{\'c}, Stephanie Samson, Manuela
  Sanguinetti, Baiba Saul{\={\i}}te, Yanin Sawanakunanon, Nathan Schneider,
  Sebastian Schuster, Djam{\'e} Seddah, Wolfgang Seeker, Mojgan Seraji,
  Mo~Shen, Atsuko Shimada, Muh Shohibussirri, Dmitry Sichinava, Natalia
  Silveira, Maria Simi, Radu Simionescu, Katalin Simk{\'o}, M{\'a}ria {\v
  S}imkov{\'a}, Kiril Simov, Aaron Smith, Isabela Soares-Bastos, Antonio
  Stella, Milan Straka, Jana Strnadov{\'a}, Alane Suhr, Umut Sulubacak, Zsolt
  Sz{\'a}nt{\'o}, Dima Taji, Yuta Takahashi, Takaaki Tanaka, Isabelle Tellier,
  Trond Trosterud, Anna Trukhina, Reut Tsarfaty, Francis Tyers, Sumire Uematsu,
  Zde{\v n}ka Ure{\v s}ov{\'a}, Larraitz Uria, Hans Uszkoreit, Sowmya Vajjala,
  Daniel van Niekerk, Gertjan van Noord, Viktor Varga, Veronika Vincze, Lars
  Wallin, Jonathan~North Washington, Seyi Williams, Mats Wir{\'e}n, Tsegay
  Woldemariam, Tak-sum Wong, Chunxiao Yan, Marat~M. Yavrumyan, Zhuoran Yu,
  Zden{\v e}k {\v Z}abokrtsk{\'y}, Amir Zeldes, Daniel Zeman, Manying Zhang,
  and Hanzhi Zhu. 2018.
\newblock {Universal Dependencies 2.2}.
\newblock {LINDAT}/{CLARIN} digital library at the {Institute of Formal and
  Applied Linguistics} ({{\'U}FAL}), {Faculty of Mathematics and Physics,
  Charles University}.

\bibitem[{Nivre et~al.(2016)Nivre, de~Marneffe, Ginter, Goldberg, Hajic,
  Manning, McDonald, Petrov, Pyysalo, Silveira et~al.}]{nivre2016universal}
Joakim Nivre, Marie-Catherine de~Marneffe, Filip Ginter, Yoav Goldberg, Jan
  Hajic, Christopher~D Manning, Ryan~T McDonald, Slav Petrov, Sampo Pyysalo,
  Natalia Silveira, et~al. 2016.
\newblock {U}niversal {D}ependencies v1: A multilingual treebank collection.
\newblock In {\em Proceedings of Language Resources and Evaluation Conference
  (LREC'16)\/}. European Language Resources Association (ELRA), Portoro{\v{z}},
  Slovenia.

\bibitem[{{\"O}stling and Bjerva(2017)}]{ostling2017surug}
Robert {\"O}stling and Johannes Bjerva. 2017.
\newblock {SU-RUG} at the {CoNLL-SIGMORPHON 2017 Shared Task}: Morphological
  inflection with attentional sequence-to-sequence models.
\newblock In {\em Proceedings of the CoNLL SIGMORPHON 2017 Shared Task:
  Universal Morphological Reinflection\/}. Association for Computational
  Linguistics, Vancouver, Canada.

\bibitem[{Pirinen(2015)}]{pirinen2015development}
Tommi~A Pirinen. 2015.
\newblock Development and use of computational morphology of {F}innish in the
  open source and open science era: Notes on experiences with {OMorFi}
  development.
\newblock {\em SKY Journal of Linguistics\/} 28:381--393.

\bibitem[{Pirinen(2019)}]{pirinen2019neural}
Tommi~A Pirinen. 2019.
\newblock Neural and rule-based {F}innish {NLP} models---expectations,
  experiments and experiences.
\newblock In {\em Proceedings of the Fifth International Workshop on
  Computational Linguistics for Uralic Languages\/}. Association for
  Computational Linguistics, Tartu, Estonia, pages 104--114.

\bibitem[{Qi et~al.(2018)Qi, Dozat, Zhang, and Manning}]{udst:stanford}
Peng Qi, Timothy Dozat, Yuhao Zhang, and Christopher~D. Manning. 2018.
\newblock {Universal Dependency} parsing from scratch.
\newblock In {\em Proceedings of the {CoNLL} 2018 Shared Task: Multilingual
  Parsing from Raw Text to Universal Dependencies\/}. Association for
  Computational Linguistics, Brussels, Belgium, pages 160--170.

\bibitem[{Sennrich et~al.(2016)Sennrich, Haddow, and
  Birch}]{sennrich2016neural}
Rico Sennrich, Barry Haddow, and Alexandra Birch. 2016.
\newblock Neural machine translation of rare words with subword units.
\newblock In {\em Proceedings of the 54rd Annual Meeting of the Association for
  Computational Linguistics (ACL 2016)\/}. Association for Computational
  Linguistics, Berlin, Germany, pages 1715--1725.

\bibitem[{Smith et~al.(2005)Smith, Smith, and Tromble}]{smith2005context}
Noah~A Smith, David~A Smith, and Roy~W Tromble. 2005.
\newblock Context-based morphological disambiguation with random fields.
\newblock In {\em Proceedings of the conference on Human Language Technology
  and Empirical Methods in Natural Language Processing\/}. Association for
  Computational Linguistics, Vancouver, Canada, pages 475--482.

\bibitem[{Straka(2018{\natexlab{a}})}]{11234/1-2859}
Milan Straka. 2018{\natexlab{a}}.
\newblock {CoNLL 2018 Shared Task} - {UDPipe} baseline models and supplementary
  materials.
\newblock {LINDAT}/{CLARIN} digital library at the Institute of Formal and
  Applied Linguistics ({{\'U}FAL}), Faculty of Mathematics and Physics, Charles
  University.

\bibitem[{Straka(2018{\natexlab{b}})}]{udst:udpipefuture}
Milan Straka. 2018{\natexlab{b}}.
\newblock {UDPipe} 2.0 prototype at {CoNLL} 2018 {UD Shared Task}.
\newblock In {\em Proceedings of the {CoNLL} 2018 Shared Task: Multilingual
  Parsing from Raw Text to Universal Dependencies\/}. Association for
  Computational Linguistics, Brussels, Belgium, pages 197--207.

\bibitem[{Straka et~al.(2016)Straka, Hajic, and
  Strakov{\'a}}]{straka2016udpipe}
Milan Straka, Jan Hajic, and Jana Strakov{\'a}. 2016.
\newblock {UDPipe}: Trainable pipeline for processing {CoNLL-U} files
  performing tokenization, morphological analysis, pos tagging and parsing.
\newblock In {\em Proceedings of the Tenth International Conference on Language
  Resources and Evaluation (LREC 2016)\/}. European Language Resources
  Association (ELRA), Portoro{\v{z}}, Slovenia.

\bibitem[{Tr{\'o}n et~al.(2006)Tr{\'o}n, Hal{\'a}csy, Rebrus, Rung, Vajda, and
  Simon}]{tron2006morphdb}
Viktor Tr{\'o}n, P{\'e}ter Hal{\'a}csy, P{\'e}ter Rebrus, Andr{\'a}s Rung,
  P{\'e}ter Vajda, and Eszter Simon. 2006.
\newblock Morphdb.hu: {H}ungarian lexical database and morphological grammar.
\newblock In {\em Proceedings of 5th International Conference on Language
  Resources and Evaluation (LREC'06)\/}. European Language Resources
  Association (ELRA), Genoa, Italy, pages 1670--1673.

\bibitem[{Tyers et~al.(2010)Tyers, S{\'a}nchez-Mart{\'\i}nez, Ortiz-Rojas, and
  Forcada}]{tyers2010free}
Francis Tyers, Felipe S{\'a}nchez-Mart{\'\i}nez, Sergio Ortiz-Rojas, and Mikel
  Forcada. 2010.
\newblock Free/open-source resources in the {A}pertium platform for machine
  translation research and development.
\newblock {\em The Prague Bulletin of Mathematical Linguistics\/} 93:67--76.

\bibitem[{Zeman et~al.(2018)Zeman, Haji{\v{c}}, Popel, Potthast, Straka,
  Ginter, Nivre, and Petrov}]{udst:overview18}
Daniel Zeman, Jan Haji{\v{c}}, Martin Popel, Martin Potthast, Milan Straka,
  Filip Ginter, Joakim Nivre, and Slav Petrov. 2018.
\newblock {CoNLL 2018 Shared Task: Multilingual Parsing from Raw Text to
  Universal Dependencies}.
\newblock In {\em Proceedings of the CoNLL 2018 Shared Task: Multilingual
  Parsing from Raw Text to Universal Dependencies\/}. Association for
  Computational Linguistics, Brussels, Belgium, pages 1--20.

\bibitem[{Zeman et~al.(2017)Zeman, Popel, Straka, Haji{\v{c}}, Nivre, Ginter,
  Luotolahti, Pyysalo, Petrov, Potthast, Tyers, Badmaeva, G{\"{o}}k{\i}rmak,
  Nedoluzhko, Cinkov{\'{a}}, Haji{\v{c}}~jr., Hlav{\'{a}}{\v{c}}ov{\'{a}},
  Kettnerov{\'{a}}, Ure{\v{s}}ov{\'{a}}, Kanerva, Ojala, Missil{\"{a}},
  Manning, Schuster, Reddy, Taji, Habash, Leung, de~Marneffe, Sanguinetti,
  Simi, Kanayama, de~Paiva, Droganova, Mart{\'{\i}}nez~Alonso, Uszkoreit,
  Macketanz, Burchardt, Harris, Marheinecke, Rehm, Kayadelen, Attia, Elkahky,
  Yu, Pitler, Lertpradit, Mandl, Kirchner, Fernandez~Alcalde, Strnadova,
  Banerjee, Manurung, Stella, Shimada, Kwak, Mendon{\c{c}}a, Lando, Nitisaroj,
  and Li}]{udst:overview17}
Daniel Zeman, Martin Popel, Milan Straka, Jan Haji{\v{c}}, Joakim Nivre, Filip
  Ginter, Juhani Luotolahti, Sampo Pyysalo, Slav Petrov, Martin Potthast,
  Francis Tyers, Elena Badmaeva, Memduh G{\"{o}}k{\i}rmak, Anna Nedoluzhko,
  Silvie Cinkov{\'{a}}, Jan Haji{\v{c}}~jr., Jaroslava
  Hlav{\'{a}}{\v{c}}ov{\'{a}}, V{\'{a}}clava Kettnerov{\'{a}}, Zde{\v{n}}ka
  Ure{\v{s}}ov{\'{a}}, Jenna Kanerva, Stina Ojala, Anna Missil{\"{a}},
  Christopher Manning, Sebastian Schuster, Siva Reddy, Dima Taji, Nizar Habash,
  Herman Leung, Marie-Catherine de~Marneffe, Manuela Sanguinetti, Maria Simi,
  Hiroshi Kanayama, Valeria de~Paiva, Kira Droganova, H{\v{e}}ctor
  Mart{\'{\i}}nez~Alonso, Hans Uszkoreit, Vivien Macketanz, Aljoscha Burchardt,
  Kim Harris, Katrin Marheinecke, Georg Rehm, Tolga Kayadelen, Mohammed Attia,
  Ali Elkahky, Zhuoran Yu, Emily Pitler, Saran Lertpradit, Michael Mandl, Jesse
  Kirchner, Hector Fernandez~Alcalde, Jana Strnadova, Esha Banerjee, Ruli
  Manurung, Antonio Stella, Atsuko Shimada, Sookyoung Kwak, Gustavo
  Mendon{\c{c}}a, Tatiana Lando, Rattima Nitisaroj, and Josie Li. 2017.
\newblock {CoNLL 2017 Shared Task: Multilingual Parsing from Raw Text to
  Universal Dependencies}.
\newblock In {\em {Proceedings of the CoNLL 2017 Shared Task: Multilingual
  Parsing from Raw Text to Universal Dependencies}\/}. Association for
  Computational Linguistics, Vancouver, Canada.

\end{thebibliography}
\bibliographystyle{acl_natbib}

\pagebreak

\appendix
\section{Lemmatization accuracy with gold segmentation and morphology}
\label{app:eval}

%%% MACHINE GENERATED TABLE %%%
{\scriptsize
\begin{longtable}{lllllllll}
%
%\centering
%\begin{tabular}{lllllllll}
\hline\hline
                     \textbf{Treebank} &       & \textbf{Tokens} & \textbf{Sents} &   \textbf{UPOS} &   \textbf{XPOS} & \textbf{UFeats} & \textbf{Lemmas} &    \textbf{LAS} \\\hline \endhead

        Afrikaans-AfriBooms &          raw text &  99.75 &     98.25 &  97.32 &  93.67 &  96.71 &  97.76 &  85.14 \\
         &          gold seg &    --- &       --- &  97.55 &  93.85 &  96.92 &  97.95 &  85.67 \\
         &   gold seg+mor &    --- &       --- &    --- &    --- &    --- &  98.63 &  88.31 \\\hline
       Ancient Greek-PROIEL &            raw text &    100.00 &     44.57 &     97.00 &  97.17 &  91.86 &  97.31 &  75.88 \\
        &            gold seg &    --- &       --- &  97.06 &  97.29 &  91.93 &  97.43 &  82.12 \\
        &     gold seg+mor &    --- &       --- &    --- &    --- &    --- &  98.87 &  84.96 \\\hline
      Ancient Greek-Perseus &           raw text &  99.96 &     98.73 &  91.93 &  83.87 &  89.94 &   89.60 &  73.26 \\
       &           gold seg &    --- &       --- &  92.01 &  83.95 &  89.99 &  89.61 &  73.42 \\
       &    gold seg+mor &    --- &       --- &    --- &    --- &    --- &  92.73 &  77.63 \\\hline
                Arabic-PADT &               raw text &  99.98 &     80.89 &  90.47 &  87.37 &  87.56 &  89.46 &  72.44 \\
                 &               gold seg &    --- &       --- &  96.75 &  93.73 &  93.95 &  95.62 &  82.46 \\
                 &        gold seg+mor &    --- &       --- &    --- &    --- &    --- &  98.34 &  84.33 \\\hline 
            Armenian-ArmTDP &             raw text &  97.21 &     92.41 &  69.31 &  96.47 &  46.05 &  71.81 &  29.74 \\
             &             gold seg &    --- &       --- &  71.21 &    100.00 &  47.71 &  73.77 &   30.90 \\
             &      gold seg+mor &    --- &       --- &    --- &    --- &    --- &  93.47 &     37.00 \\\hline
                 Basque-BDT &                raw text &  99.96 &     99.08 &  96.01 &  99.96 &  92.07 &  96.81 &  82.38 \\
                  &                gold seg &    --- &       --- &  96.06 &    100.00 &   92.10 &  96.86 &  82.52 \\
                  &         gold seg+mor &    --- &       --- &    --- &    --- &    --- &   98.50 &  86.08 \\\hline
              Bulgarian-BTB &                raw text &  99.92 &     92.85 &  98.61 &  96.41 &  97.41 &  98.17 &   89.60 \\
               &                gold seg &    --- &       --- &   98.70 &   96.50 &   97.50 &  98.25 &  90.54 \\
               &         gold seg+mor &    --- &       --- &    --- &    --- &    --- &  99.42 &  91.59 \\\hline
                 Buryat-BDT &               raw text &  97.07 &      90.90 &  37.95 &  97.07 &  35.42 &  56.05 &  13.13 \\
                  &               gold seg &    --- &       --- &  39.32 &    100.00 &  37.24 &  58.11 &  13.43 \\
                  &        gold seg+mor &    --- &       --- &    --- &    --- &    --- &  90.99 &  15.35 \\\hline
             Catalan-AnCora &             raw text &  99.97 &     99.03 &  97.32 &  97.34 &  96.66 &  97.66 &  90.04 \\
              &             gold seg &    --- &       --- &  97.36 &  97.37 &   96.70 &  97.69 &  90.15 \\
              &      gold seg+mor &    --- &       --- &    --- &    --- &    --- &  99.24 &   92.30 \\\hline
                Chinese-GSD &                raw text &  89.55 &      98.20 &  85.83 &   85.60 &  88.74 &  89.55 &  66.26 \\
                 &                gold seg &    --- &       --- &  95.28 &     95.00 &   99.10 &    100.00 &  81.38 \\
                 &         gold seg+mor &    --- &       --- &    --- &    --- &    --- &    100.00 &  86.61 \\\hline
               Croatian-SET &                raw text &  99.92 &     95.36 &  98.02 &  99.92 &  91.91 &  96.87 &  86.19 \\
                &                gold seg &    --- &       --- &  98.09 &    100.00 &  91.94 &  96.94 &  86.65 \\
                &         gold seg+mor &    --- &       --- &    --- &    --- &    --- &  98.37 &  87.78 \\\hline
                  Czech-CAC &                raw text &  99.97 &     99.76 &  98.96 &  95.37 &  94.65 &  98.34 &  90.52 \\
                   &                gold seg &    --- &       --- &     99.00 &  95.41 &  94.69 &  98.39 &  90.59 \\
                   &         gold seg+mor &    --- &       --- &    --- &    --- &    --- &  99.48 &  91.72 \\\hline
              Czech-FicTree &            raw text &  99.97 &     98.37 &  98.51 &  94.56 &  95.29 &  98.84 &  90.63 \\
               &            gold seg &    --- &       --- &  98.53 &  94.59 &  95.33 &  98.87 &  90.83 \\
               &     gold seg+mor &    --- &       --- &    --- &    --- &    --- &  99.63 &  92.39 \\\hline
                  Czech-PDT &                raw text &  99.93 &     92.29 &  98.72 &  95.41 &  95.18 &  98.52 &  90.54 \\
                   &                gold seg &    --- &       --- &   98.80 &  95.54 &   95.30 &   98.60 &  91.48 \\
                   &         gold seg+mor &    --- &       --- &    --- &    --- &    --- &  99.68 &  92.42 \\\hline
                  Czech-PUD &                raw text &  99.28 &      95.40 &   96.10 &  92.19 &  92.11 &  96.14 &  84.71 \\
                   &                gold seg &    --- &       --- &  96.51 &  92.62 &  92.57 &  96.54 &  85.58 \\
                   &         gold seg+mor &    --- &       --- &    --- &    --- &    --- &  97.52 &   87.20 \\\hline
                 Danish-DDT &                raw text &  99.87 &     87.96 &  97.31 &  99.87 &   97.10 &  97.88 &  82.96 \\
                  &                gold seg &    --- &       --- &  97.47 &    100.00 &  97.25 &  98.01 &  84.34 \\
                  &         gold seg+mor &    --- &       --- &    --- &    --- &    --- &   99.20 &  86.75 \\\hline
               Dutch-Alpino &             raw text &  99.83 &      90.80 &  96.13 &  94.44 &  96.44 &  96.85 &  85.36 \\
                &             gold seg &    --- &       --- &  96.32 &  94.65 &  96.61 &  97.04 &  86.56 \\
                &      gold seg+mor &    --- &       --- &    --- &    --- &    --- &  98.61 &  89.41 \\\hline
           Dutch-LassySmall &         raw text &  99.82 &     72.23 &  95.87 &  94.22 &  95.66 &  97.44 &  81.31 \\
            &         gold seg &    --- &       --- &  96.08 &  94.59 &  96.03 &  97.68 &  84.95 \\
            &  gold seg+mor &    --- &       --- &    --- &    --- &    --- &  99.39 &  88.42 \\\hline
                English-EWT &                raw text &  99.03 &     75.33 &  94.85 &  94.64 &  95.95 &  96.94 &  82.67 \\
                 &                gold seg &    --- &       --- &  95.77 &  95.63 &   96.90 &  97.81 &  86.93 \\
                 &         gold seg+mor &    --- &       --- &    --- &    --- &    --- &  99.73 &  89.57 \\\hline
                English-GUM &                raw text &  99.75 &     78.79 &  93.37 &  93.28 &  95.69 &  96.21 &  80.67 \\
                 &                gold seg &    --- &       --- &  93.64 &  93.56 &  95.99 &  96.39 &     83.00 \\
                 &         gold seg+mor &    --- &       --- &    --- &    --- &    --- &  98.08 &     86.00 \\\hline
              English-LinES &              raw text &  99.95 &     88.08 &  96.43 &  95.04 &  96.74 &  96.79 &  79.44 \\
               &              gold seg &    --- &       --- &  96.49 &  95.13 &  96.79 &  96.83 &  80.09 \\
               &       gold seg+mor &    --- &       --- &    --- &    --- &    --- &  97.19 &  81.66 \\\hline
                English-PUD &                raw text &  99.74 &     95.57 &   94.90 &   94.20 &  95.07 &   96.40 &   85.40 \\
                 &                gold seg &    --- &       --- &  95.13 &  94.44 &  95.33 &  96.65 &  86.06 \\
                 &         gold seg+mor &    --- &       --- &    --- &    --- &    --- &  98.94 &  88.23 \\\hline \pagebreak
               Estonian-EDT &                raw text &  99.91 &     90.02 &  96.43 &  97.87 &  95.71 &   96.60 &  84.14 \\
                &                gold seg &    --- &       --- &  96.49 &  97.94 &  95.78 &  96.69 &  85.09 \\
                &         gold seg+mor &    --- &       --- &    --- &    --- &    --- &  98.34 &  87.92 \\\hline
                Finnish-FTB &                raw text &    100.00 &     87.04 &  96.18 &  95.15 &  96.45 &  97.18 &  86.99 \\
                 &                gold seg &    --- &       --- &  96.29 &  95.23 &  96.55 &  97.22 &  88.92 \\
                 &         gold seg+mor &    --- &       --- &    --- &    --- &    --- &  99.19 &  92.34 \\\hline
                Finnish-PUD &                raw text &  99.63 &      92.20 &  96.91 &      0.00 &  96.72 &  95.13 &  88.88 \\
                 &                gold seg &    --- &       --- &  97.23 &      0.00 &  97.03 &  95.43 &  89.23 \\
                 &         gold seg+mor &    --- &       --- &    --- &    --- &    --- &  96.67 &  88.92 \\\hline
                Finnish-TDT &                raw text &  99.69 &     86.75 &  96.57 &  97.61 &  95.41 &   95.40 &  86.48 \\
                 &                gold seg &    --- &       --- &  96.92 &  97.89 &  95.69 &   95.70 &  88.37 \\
                 &         gold seg+mor &    --- &       --- &    --- &    --- &    --- &  97.68 &  90.84 \\\hline\hline

                 French-GSD &                raw text &  99.66 &     92.12 &  95.96 &  98.78 &  95.73 &  96.91 &  85.62 \\
                  &                gold seg &    --- &       --- &  97.15 &    100.00 &  96.86 &  98.11 &  87.58 \\
                  &         gold seg+mor &    --- &       --- &    --- &    --- &    --- &  99.38 &  88.81 \\\hline 
             French-Sequoia &            raw text &  99.79 &     82.77 &  97.42 &  99.09 &  97.04 &  98.06 &  87.42 \\
              &           gold seg &    --- &       --- &  98.44 &    100.00 &  97.97 &     99.00 &  89.96 \\
              &     gold seg+mor &    --- &       --- &    --- &    --- &    --- &  99.66 &  91.19 \\\hline
              French-Spoken &             raw text &    100.00 &     21.63 &  94.72 &  97.51 &    100.00 &  97.04 &  69.15 \\
               &             gold seg &    --- &       --- &  94.81 &  97.49 &    100.00 &  97.21 &  76.18 \\
               &      gold seg+mor &    --- &       --- &    --- &    --- &    --- &  97.95 &  78.58 \\\hline
               Galician-CTG &                raw text &  99.84 &     96.59 &  97.07 &  96.87 &  98.96 &  97.92 &  81.64 \\
                &                gold seg &    --- &       --- &  97.85 &  97.67 &  99.78 &  98.73 &  83.25 \\
                &         gold seg+mor &    --- &       --- &    --- &    --- &    --- &  99.42 &  85.22 \\\hline
           Galician-TreeGal &            raw text &  99.69 &      83.90 &  93.81 &  90.97 &  92.92 &   95.50 &  72.88 \\
            &            gold seg &    --- &       --- &  94.95 &  91.92 &  93.96 &  96.53 &  76.04 \\
            &     gold seg+mor &    --- &       --- &    --- &    --- &    --- &  98.02 &  79.42 \\\hline
                 German-GSD &                raw text &  99.58 &     81.32 &  93.81 &  96.56 &   90.20 &  96.56 &  78.64 \\
                  &               gold seg &    --- &       --- &  94.11 &  97.02 &  90.87 &  96.95 &  80.72 \\
                  &         gold seg+mor &    --- &       --- &    --- &    --- &    --- &  97.28 &  82.82 \\\hline
              Gothic-PROIEL &            raw text &    100.00 &     28.03 &  95.49 &  96.14 &  89.07 &  96.21 &   67.70 \\
               &            gold seg &    --- &       --- &  96.31 &  96.67 &  89.75 &  96.29 &  78.08 \\
               &     gold seg+mor &    --- &       --- &    --- &    --- &    --- &  98.29 &  82.01 \\\hline
                  Greek-GDT &                raw text &  99.86 &     90.11 &  97.62 &  97.52 &  94.18 &  97.26 &  88.21 \\
                   &                gold seg &    --- &       --- &  97.79 &  97.72 &  94.41 &  97.38 &  89.09 \\
                   &         gold seg+mor &    --- &       --- &    --- &    --- &    --- &  98.17 &  90.17 \\\hline
                 Hebrew-HTB &                raw text &  99.98 &       100.00 &   82.70 &  82.71 &  81.05 &  82.93 &  64.47 \\
                  &                gold seg &    --- &       --- &  97.23 &  97.28 &  95.59 &  97.19 &  85.76 \\
                  &         gold seg+mor &    --- &       --- &    --- &    --- &    --- &   98.50 &  87.79 \\\hline
                 Hindi-HDTB &               raw text &    100.00 &      99.20 &  97.43 &  96.93 &  93.91 &   98.70 &  91.58 \\
                  &               gold seg &    --- &       --- &  97.44 &  96.93 &  93.91 &   98.70 &  91.63 \\
                  &        gold seg+mor &    --- &       --- &    --- &    --- &    --- &  98.91 &  94.02 \\\hline
           Hungarian-Szeged &             raw text &  99.81 &     95.58 &  94.08 &  99.81 &  92.47 &  94.57 &  78.53 \\
            &             gold seg &    --- &       --- &   94.20 &    100.00 &  92.63 &   94.70 &  79.04 \\
            &      gold seg+mor &    --- &       --- &    --- &    --- &    --- &  97.45 &  83.65 \\\hline
             Indonesian-GSD &                raw text &    100.00 &        92.00 &  91.93 &  94.52 &  95.59 &  99.68 &  78.34 \\
              &                gold seg &    --- &       --- &  91.94 &  94.52 &  95.61 &  99.68 &  78.65 \\
              &         gold seg+mor &    --- &       --- &    --- &    --- &    --- &  99.85 &   81.40 \\\hline
                  Irish-IDT &                raw text &   99.30 &      92.60 &  92.36 &  91.05 &  82.47 &  90.52 &  70.88 \\
                   &                gold seg &    --- &       --- &  93.02 &  91.68 &  83.16 &  91.15 &   71.80 \\
                   &         gold seg+mor &    --- &       --- &    --- &    --- &    --- &  94.47 &  74.47 \\\hline
               Italian-ISDT &               raw text &  99.75 &     96.81 &  97.63 &  97.41 &  97.51 &  98.16 &  90.22 \\
                &               gold seg &    --- &       --- &  97.94 &  97.73 &  97.77 &  98.44 &  90.91 \\
                &        gold seg+mor &    --- &       --- &    --- &    --- &    --- &  99.35 &   92.70 \\\hline
           Italian-PoSTWITA &           raw text &  99.73 &      21.80 &  95.71 &  95.38 &  95.76 &  96.63 &  72.22 \\
            &           gold seg &    --- &       --- &  96.29 &  95.97 &  96.31 &  97.17 &   81.40 \\
            &    gold seg+mor &    --- &       --- &    --- &    --- &    --- &  98.79 &  83.61 \\\hline
               Japanese-GSD &                raw text &  90.46 &     95.01 &  88.84 &  90.46 &  90.45 &  89.63 &  74.52 \\
                &                gold seg &    --- &       --- &  97.84 &    100.00 &  99.98 &  98.78 &  92.43 \\
                &         gold seg+mor &    --- &       --- &    --- &    --- &    --- &  99.13 &  94.16 \\\hline
                 Kazakh-KTB &                raw text &  93.11 &     81.56 &  51.06 &  46.83 &   35.10 &  57.43 &  22.79 \\
                  &                gold seg &    --- &       --- &  55.38 &  51.19 &  37.54 &  61.83 &   26.50 \\
                  &         gold seg+mor &    --- &       --- &    --- &    --- &    --- &  95.66 &   33.70 \\\hline
                 Korean-GSD &                raw text &  99.81 &     90.49 &  96.09 &  90.53 &  99.59 &  93.94 &  83.46 \\
                  &                gold seg &    --- &       --- &  96.35 &  90.76 &  99.79 &  94.11 &  84.68 \\
                  &         gold seg+mor &    --- &       --- &    --- &    --- &    --- &  97.94 &  86.69 \\\hline
               Korean-Kaist &              raw text &    100.00 &       100.00 &  95.61 &  87.15 &    100.00 &  94.39 &  86.77 \\
                &              gold seg &    --- &       --- &  95.61 &  87.15 &    100.00 &  94.39 &  86.77 \\
                &       gold seg+mor &    --- &       --- &    --- &    --- &    --- &  98.85 &  89.91 \\\hline \pagebreak
                Kurmanji-MG &                raw text &  94.33 &     69.14 &  55.42 &  51.87 &  42.01 &  64.83 &  23.44 \\
                 &                gold seg &    --- &       --- &  57.28 &  52.78 &  43.51 &  67.94 &  25.19 \\
                 &         gold seg+mor &    --- &       --- &    --- &    --- &    --- &  91.72 &  28.88 \\\hline
                 Latin-ITTB &               raw text &  99.94 &     82.49 &  98.32 &  94.47 &  95.35 &  98.67 &  86.53 \\
                  &               gold seg &    --- &       --- &  98.38 &  94.56 &  95.44 &   98.70 &  88.97 \\
                  &        gold seg+mor &    --- &       --- &    --- &    --- &    --- &  99.64 &  91.04 \\\hline
               Latin-PROIEL &             raw text &  99.99 &     35.16 &  96.61 &  96.81 &  90.94 &   97.20 &  71.47 \\
                &             gold seg &    --- &       --- &  96.87 &  97.05 &  91.51 &  97.27 &  80.54 \\
                &      gold seg+mor &    --- &       --- &    --- &    --- &    --- &  98.97 &  83.73 \\\hline
              Latin-Perseus &            raw text &    100.00 &     98.35 &  90.52 &  75.01 &  79.18 &  85.27 &  62.27 \\
               &            gold seg &    --- &       --- &  90.49 &     75.00 &  79.15 &  85.26 &  62.42 \\
               &     gold seg+mor &    --- &       --- &    --- &    --- &    --- &  91.67 &  69.58 \\\hline
               Latvian-LVTB &               raw text &   99.40 &     98.34 &  94.99 &  86.46 &  91.08 &  93.95 &  80.83 \\
                &               gold seg &    --- &       --- &  95.47 &  86.88 &  91.57 &  94.45 &  81.67 \\
                &        gold seg+mor &    --- &       --- &    --- &    --- &    --- &  98.53 &   85.30 \\\hline 
          North Sami-Giella &            raw text &  99.84 &     98.33 &  91.37 &  92.94 &  88.11 &   89.70 &   69.60 \\
           &            gold seg &    --- &       --- &  91.49 &  93.17 &  88.33 &  89.83 &  69.89 \\
           &     gold seg+mor &    --- &       --- &    --- &    --- &    --- &  94.71 &  80.11 \\\hline

          Norwegian-Bokmaal &            raw text &  99.78 &     95.79 &  97.33 &  99.78 &  96.25 &  97.97 &  89.52 \\
           &            gold seg &    --- &       --- &  97.53 &    100.00 &  96.45 &  98.18 &  90.27 \\
           &     gold seg+mor &    --- &       --- &    --- &    --- &    --- &  99.62 &  93.12 \\\hline
          Norwegian-Nynorsk &            raw text &  99.93 &     92.08 &  97.18 &  99.93 &  96.25 &  97.72 &  89.46 \\
           &            gold seg &    --- &       --- &   97.30 &    100.00 &  96.38 &  97.81 &  90.31 \\
           &     gold seg+mor &    --- &       --- &    --- &    --- &    --- &  99.41 &  92.82 \\\hline
       Norwegian-NynorskLIA &         raw text &  99.99 &     99.86 &  89.56 &  99.99 &  88.84 &  94.51 &  57.45 \\
        &         gold seg &    --- &       --- &  89.63 &    100.00 &  88.71 &  94.53 &  57.37 \\
        &  gold seg+mor &    --- &       --- &    --- &    --- &    --- &   98.50 &  65.13 \\\hline
 Old Church Slavonic-PROIEL &             raw text &    100.00 &     37.28 &  96.09 &   96.20 &  89.39 &  95.14 &  73.36 \\
  &             gold seg &    --- &       --- &  96.35 &  96.53 &  89.76 &  95.27 &  83.43 \\
  &      gold seg+mor &    --- &       --- &    --- &    --- &    --- &  97.29 &  87.49 \\\hline
             Persian-Seraji &             raw text &    100.00 &     98.74 &  97.02 &  97.05 &  97.12 &  96.77 &  85.98 \\
             &             gold seg &    --- &       --- &  97.31 &  97.34 &  97.41 &  97.03 &  86.53 \\
              &      gold seg+mor &    --- &       --- &    --- &    --- &    --- &  97.29 &  89.62 \\\hline
                 Polish-LFG &                raw text &  99.86 &     99.74 &  98.26 &  93.54 &  94.57 &  97.66 &  94.51 \\
                  &                gold seg &    --- &       --- &  98.39 &  93.65 &  94.69 &  97.77 &  94.89 \\
                  &         gold seg+mor &    --- &       --- &    --- &    --- &    --- &  99.47 &  96.64 \\\hline
                  Polish-SZ &                 raw text &  99.99 &        99.00 &  97.85 &  92.03 &  92.13 &  97.08 &  91.15 \\
                   &                 gold seg &    --- &       --- &  97.99 &   92.20 &   92.30 &  97.21 &  91.76 \\
                   &          gold seg+mor &    --- &       --- &    --- &    --- &    --- &  98.86 &   94.40 \\\hline
          Portuguese-Bosque &             raw text &  99.71 &     88.79 &  96.07 &  99.59 &  95.73 &  97.58 &  87.42 \\
           &             gold seg &    --- &       --- &  96.43 &    100.00 &  96.11 &  98.02 &  88.56 \\
           &      gold seg+mor &    --- &       --- &    --- &    --- &    --- &   98.70 &  89.42 \\\hline
               Romanian-RRT &                raw text &  99.67 &     93.72 &  97.53 &  97.04 &   97.20 &  98.23 &  86.04 \\
                &                gold seg &    --- &       --- &  97.86 &  97.35 &  97.51 &  98.54 &  86.87 \\
                &         gold seg+mor &    --- &       --- &    --- &    --- &    --- &  99.63 &  87.53 \\\hline
          Russian-SynTagRus &          raw text &   99.60 &     98.01 &  97.98 &   99.60 &  96.46 &  98.15 &  91.66 \\
           &          gold seg &    --- &       --- &  98.37 &    100.00 &  96.85 &  98.53 &  92.51 \\
           &   gold seg+mor &    --- &       --- &    --- &    --- &    --- &   99.60 &  93.63 \\\hline
              Russian-Taiga &              raw text &  98.14 &     87.38 &  91.88 &  98.12 &  82.23 &  89.32 &  64.15 \\
               &              gold seg &    --- &       --- &  93.33 &  99.98 &  83.86 &   90.80 &  66.67 \\
               &       gold seg+mor &    --- &       --- &    --- &    --- &    --- &   96.40 &  69.79 \\\hline
                Serbian-SET &                raw text &  99.97 &     92.02 &  97.97 &  99.97 &  93.95 &  97.17 &   88.60 \\
                 &                gold seg &    --- &       --- &  97.99 &    100.00 &     94.00 &   97.20 &  89.11 \\
                 &         gold seg+mor &    --- &       --- &    --- &    --- &    --- &  98.68 &  90.37 \\\hline
                 Slovak-SNK &                raw text &    100.00 &     84.26 &  95.69 &   84.30 &  89.94 &  96.35 &  86.61 \\
                  &                gold seg &    --- &       --- &  95.69 &  84.44 &  90.06 &  96.35 &  88.43 \\
                  &         gold seg+mor &    --- &       --- &    --- &    --- &    --- &   98.30 &  90.72 \\\hline
              Slovenian-SSJ &                raw text &  98.29 &     76.61 &  96.49 &  92.34 &  92.82 &  96.49 &  86.78 \\
               &                gold seg &    --- &       --- &  98.15 &  94.21 &  94.73 &  98.21 &  91.63 \\
               &         gold seg+mor &    --- &       --- &    --- &    --- &    --- &  99.49 &  94.22 \\\hline
              Slovenian-SST &                raw text &    100.00 &      22.90 &  93.87 &  85.43 &  85.39 &   94.90 &  53.97 \\
               &                gold seg &    --- &       --- &  94.35 &  85.45 &  85.54 &  95.04 &  66.26 \\
               &         gold seg+mor &    --- &       --- &    --- &    --- &    --- &  98.89 &  72.45 \\\hline
             Spanish-AnCora &             raw text &  99.97 &     98.26 &   97.80 &  97.76 &  97.34 &  98.48 &  89.62 \\
              &             gold seg &    --- &       --- &  97.85 &  97.81 &  97.39 &  98.52 &  89.87 \\
              &      gold seg+mor &    --- &       --- &    --- &    --- &    --- &  99.71 &  91.78 \\\hline
              Swedish-LinES &              raw text &  99.96 &     85.25 &  96.63 &  94.58 &  89.55 &  97.29 &  81.16 \\
               &              gold seg &    --- &       --- &  96.64 &  94.63 &  89.63 &  97.34 &  82.23 \\
               &       gold seg+mor &    --- &       --- &    --- &    --- &    --- &  98.41 &  84.06 \\\hline \pagebreak
                Swedish-PUD &                raw text &  98.41 &     94.47 &  93.58 &  91.88 &  77.99 &  87.47 &  79.15 \\
                 &                gold seg &    --- &       --- &  94.31 &  92.48 &   78.80 &  88.86 &  80.42 \\
                 &         gold seg+mor &    --- &       --- &    --- &    --- &    --- &  90.42 &   82.90 \\\hline
          Swedish-Talbanken &          raw text &  99.78 &     93.17 &  97.47 &  96.41 &  96.63 &  97.98 &  85.87 \\
           &          gold seg &    --- &       --- &  97.64 &  96.55 &  96.77 &  98.15 &  86.74 \\
           &   gold seg+mor &    --- &       --- &    --- &    --- &    --- &   98.70 &   89.10 \\\hline
               Turkish-IMST &               raw text &  99.86 &     97.09 &  94.32 &  93.27 &  91.05 &  95.16 &   64.70 \\
                &               gold seg &    --- &       --- &  96.19 &  95.02 &  92.83 &  97.03 &  67.84 \\
                &        gold seg+mor &    --- &       --- &    --- &    --- &    --- &  99.42 &  69.95 \\\hline
               Ukrainian-IU &                 raw text &  99.67 &     95.04 &  97.04 &  90.71 &  90.83 &  96.62 &  84.43 \\
                &                 gold seg &    --- &       --- &  97.41 &  90.96 &  91.09 &  96.93 &  85.24 \\
                &          gold seg+mor &    --- &       --- &    --- &    --- &    --- &  99.16 &     88.00 \\\hline
         Upper Sorbian-UFAL &              raw text &   98.60 &     74.51 &  59.51 &   98.60 &  39.66 &   54.80 &   24.90 \\
          &              gold seg &    --- &       --- &  60.42 &    100.00 &  40.58 &  55.26 &  26.13 \\
          &       gold seg+mor &    --- &       --- &    --- &    --- &    --- &  73.74 &  34.51 \\\hline 
                  Urdu-UDTB &               raw text &    100.00 &      98.60 &  94.54 &  92.74 &   83.90 &  97.43 &  82.15 \\
                   &               gold seg &    --- &       --- &  94.54 &  92.73 &  83.91 &  97.43 &   82.20 \\
                   &        gold seg+mor &    --- &       --- &    --- &    --- &    --- &  97.99 &  86.85 \\\hline
                 Uyghur-UDT &                raw text &  99.22 &     81.61 &  89.15 &  91.56 &  87.45 &  94.15 &  62.92 \\
                  &                gold seg &    --- &       --- &  89.96 &  92.24 &  88.16 &  94.95 &  65.07 \\
                  &         gold seg+mor &    --- &       --- &    --- &    --- &    --- &  99.01 &  68.44 \\\hline
             Vietnamese-VTB &                raw text &  84.26 &     92.87 &   76.50 &  73.79 &  83.93 &  84.26 &  42.87 \\
              &                gold seg &    --- &       --- &  89.15 &  85.36 &  99.53 &  99.98 &  59.88 \\
              &         gold seg+mor &    --- &       --- &    --- &    --- &    --- &  99.98 &  69.38 \\\hline

%\end{tabular}
\\
    \caption{Lemmatization accuracy for all treebanks measured on gold and predicted segmentation and tagging.}
    \label{tbl:appeval}
\end{longtable}}

%%% MACHINE GENERATED TABLE %%%

\label{lastpage}

\end{document}